\pdfoutput=1
\PassOptionsToPackage{dvipsnames}{xcolor}
\documentclass[11pt]{article}

\usepackage[preprint]{acl}

\usepackage{times}
\usepackage{latexsym}

\usepackage[T1]{fontenc}

\usepackage[utf8]{inputenc}

\usepackage{inconsolata}

\usepackage{microtype}
\usepackage{hyperref}
\usepackage{url}
\usepackage{booktabs}

\usepackage{bbding}
\usepackage{hyperref}
\usepackage{url}
\usepackage{subcaption}
\usepackage{enumitem}
\usepackage{graphicx}
\usepackage{comment}
\usepackage{multicol}
\usepackage{mdwlist}
\usepackage{multirow}
\usepackage{booktabs}
\usepackage{wrapfig}
\usepackage{amsmath,amssymb,bm}
\usepackage[most]{tcolorbox}
\usepackage{cuted,lipsum}
\tcbuselibrary{skins, breakable, theorems,fitting}
\usepackage{caption}

\newtcbox{\mybox}[1][red]
  {on line, arc = 0pt, outer arc = 0pt,
    colback = #1!10!white, colframe = #1!50!black,
    boxsep = 0pt, left = 1pt, right = 1pt, top = 2pt, bottom = 2pt,
    boxrule = 0pt, bottomrule = 1pt, toprule = 1pt}

\title{Skills-in-Context:  Unlocking Compositionality in Large Language Models}

\author{\normalsize Jiaao Chen\thanks{Affiliated with Georgia Institute of Technology. This work is done during internship at Tencent AI Lab.}, Xiaoman Pan, Dian Yu, Kaiqiang Song, Xiaoyang Wang, Dong Yu \& Jianshu Chen\thanks{Corresponding to: Jiaao Chen - \texttt{jiaaochen@gatech.edu}; Jianshu Chen - \texttt{jianshuchen@global.tencent.com}.} \\
\normalsize Tencent AI Lab, Bellevue, WA, 98004
}

\begin{document}
\maketitle
\begin{abstract}
We investigate how to elicit compositional generalization capabilities in large language models (LLMs). Compositional generalization empowers LLMs to solve complex problems by combining foundational skills, a critical reasoning ability akin to human intelligence.
However, even the most advanced LLMs currently struggle with this form of reasoning.
We examine this problem within the framework of in-context learning and find that demonstrating both foundational skills and compositional examples grounded in these skills within the same prompt context is crucial. We refer to this prompt structure as \emph{skills-in-context} (SKiC). With as few as two exemplars, this in-context learning structure enables LLMs to tackle more challenging problems requiring innovative skill combinations, achieving near-perfect systematic generalization across a broad range of tasks. Intriguingly, SKiC also unlocks the latent potential of LLMs, allowing them to more actively utilize pre-existing internal skills acquired during earlier pretraining stages to solve complex reasoning problems. The SKiC structure is robust across different skill constructions and exemplar choices and demonstrates strong transferability to new tasks. Finally, inspired by our in-context learning study, we show that fine-tuning LLMs with SKiC-style data can elicit zero-shot weak-to-strong generalization, enabling the models to solve much harder problems directly with standard prompting.
\end{abstract}

\section{Introduction}
\label{sec: introduction}

Large language models (LLMs) have achieved great success in solving natural language processing (NLP) tasks \citep{smith2022using, lewkowycz2022solving, wei2021finetuned, mishra2022cross, chung2022scaling, ouyang2022training,openai2023gpt4,touvron2023llama}. When the size of model and data scales up, LLMs exhibit strong zero/few-shot performance on a wide range of NLP tasks --- a salient behavior characterized by the scaling law \citep{kaplan2020scaling,hoffmann2022training} and emergent abilities \citep{wei2022emergent}. However, LLMs still struggle with compositional generalization, i.e., the ability to use existing skills to solve more complex unseen problems~\citep{zhou2022least,dziri2023faith,burns2023weak}. 


\begin{figure*}[t]
\begin{center}
\centerline{\includegraphics[width=2.0\columnwidth]{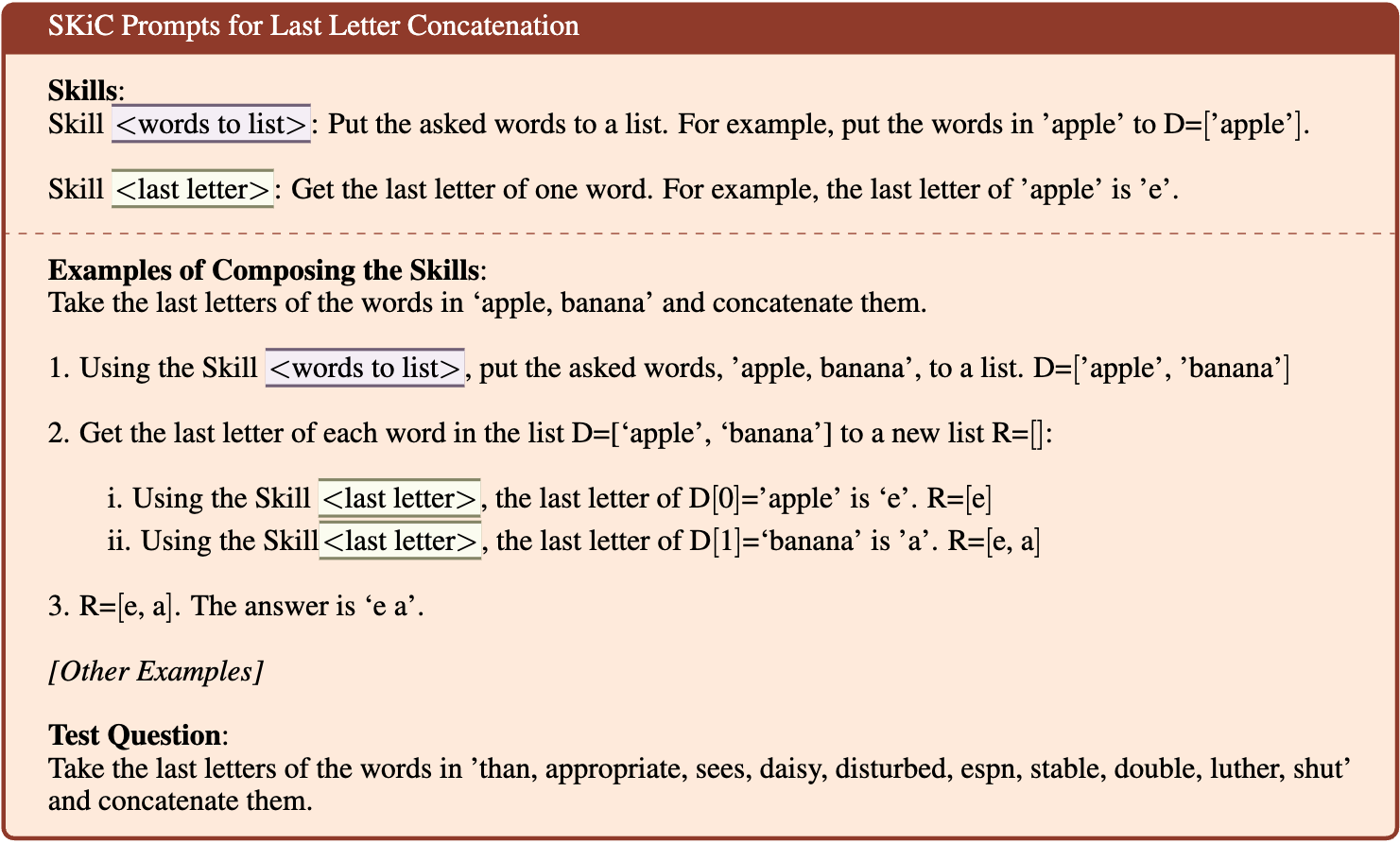}}
\caption{Skills-in-Context Prompting. The prompt consists of three blocks: (i) the (basic) skills for solving a complex task, (ii) examples of how to compose the skills, and (iii) the problem to be solved. The above prompt will be fed into an LLM to generate the output --- see Figure \ref{Tab:example_last_letter_skill} for an example of the output. Note that the compositional exemplars demonstrate how to \emph{explicitly} ground the reasoning steps onto the basic skills (highlighted in colors).
} \label{fig:example}
\end{center}
\vskip -0.2in
\end{figure*}

Ideally, if an LLM has already learned a rich set of knowledge and foundational skills, it should be able to solve any problem whose solutions are composable from these skills. To unlock such great potential, the key is to teach the LLMs how to use these skills to construct a solution to more difficult problems. Towards this goal, there have been a series of in-context learning strategies developed to improve the reasoning and composition capabilities. Notably, chain-of-thought (CoT) prompting \citep{wei2022chain} significantly improves the reasoning performance of LLMs by demonstrating how to approach a complex problem through a sequence of basic steps. Follow-ups such as Least-to-Most prompting \citep{zhou2022least} and decomposed prompting \citep{khot2022decomposed} propose a two-stage strategy, which first decomposes the problem into sub-problems, and then solve and combine them sequentially. Although these methods significantly boost the performance in solving many challenging compositional generalization tasks, they usually fail over problems that are significantly harder than the ones they have seen. Moreover, least-to-most prompting and decomposed prompting are restricted to solving problem classes that can be decomposed as a sequence of sub-problems. And for problems with general computation graphs \citep{dziri2023faith}, it is generally less intuitive, if not possible, to construct the prompting exemplars.

In this paper, we examine how to elicit strong compositional abilities in LLMs within the framework of in-context learning. We find that the key insight is to teach the LLM to \textbf{explicitly ground each of its reasoning steps on the (more foundational) skills}. To this end, it is crucial to demonstrate both the foundational skills and the compositional examples grounded in these skills within the same prompt context. We refer to this (\emph{one-stage}) prompting structure as \textbf{SK}ills-\textbf{i}n-\textbf{C}ontext (SKiC).
Specifically, the SKiC prompt is constructed from three main blocks (Figure \ref{fig:example}). The first block contains a short (non-exhaustive) list of skills that LLMs may need to use in order to solve a more complex problem, which include the instructions of the skills. These skills can be distilled either manually or automatically via LLMs. The second part consists of a few (generally two) exemplars that demonstrate how to compose skills into a complex solution. The last part is the testing problem. 

Interestingly, with both the skills and their explicit compositions presented in the context, the LLMs successfully learn how to ground reasoning steps onto the skills that they have already mastered, yielding much stronger generalization abilities. It allows LLMs to achieve near-perfect systematic generalization across a broad range of tasks. In addition, it also allows the LLMs to generalize beyond the skills provided in the context and solve problems by more actively and explicitly using the vast reservoir of the internal skills they acquired during the prior pre-training stage. It clearly demonstrates that SKiC structure unleashes strong synergies between skills and their composition capabilities, which teaches LLMs to generalize to unseen (harder) problems that require innovative compositions of skills. Furthermore, the SKiC structure is robust across different skill constructions (e.g., handcrafted or discovered by LLMs) and exemplar choices and demonstrates strong transferability to new tasks. Finally, inspired by our in-context learning study, we show that fine-tuning LLMs with SKiC-style data can elicit zero-shot weak-to-strong generalization, enabling the models to solve much harder problems directly with standard prompting.



\section{SKiC: Elicit Compositionality with In-Context Skills and Grounding}
\label{sec: methodology}
While humans naturally exhibit compositional generalization in problem-solving, LLMs often struggle to compose \textit{basic skills} to solve more difficult problems \citep{dziri2023faith}. Empowering LLMs with the ability to compose skills that they have seen to solve more complex tasks is important to mirror human intelligence and to reach superintelligence. In this work, we investigate how to elicit compositionality of LLMs in in-context learning (ICL) setting. In particular, we want to reveal how a meticulously designed prompt structure could greatly enhance the compositional ability. The insights obtained in the ICL setting can also inspire how to further improve the fine-tuning (Sec. \ref{sec:ift}).

\paragraph{Demonstration of Composition}
We find that it is crucial to instruct the LLM to explicitly ground each of its reasoning steps onto the foundational skills\footnote{``Foundational skills'' are not necessarily atomic. Rather, they could be any skills (e.g., a composite skill by itself) that serve as the building blocks for tackling complex problems.}. To facilitate this, it is important to demonstrate both the foundational skills and the compositional examples grounded in these skills within the same prompt context. Such a structure, which we refer to as SKiC, provides a full-context demonstration of how to perform \emph{explicit} composition over skills for solving a (complex) problem, where the detailed three-part construction is illustrated in Figure \ref{fig:example} as we discussed earlier. It is also partly inspired by the Elaborative Rehearsal from the human cognition theory \citep{berry1983metacognitive}, where studies \citep{kheirzadeh2016depth} have demonstrated that by first summarizing relevant knowledge and skills as the Scaffolding \citep{hammond2005putting} and establishing connections between the problem-solving steps and the existing Scaffolding, human would process the new information with greater depth and thoroughness, thus reinforcing both the concepts and their practical applications \citep{bakker2015scaffolding}. Our ablation study in Table \ref{Tab:different_components_main} will reveal that both the in-context skills and the explicit groundings are essential for eliciting strong compositional abilities.

\begin{figure*}[t]
\begin{center}
\centerline{\includegraphics[width=2.0\columnwidth]{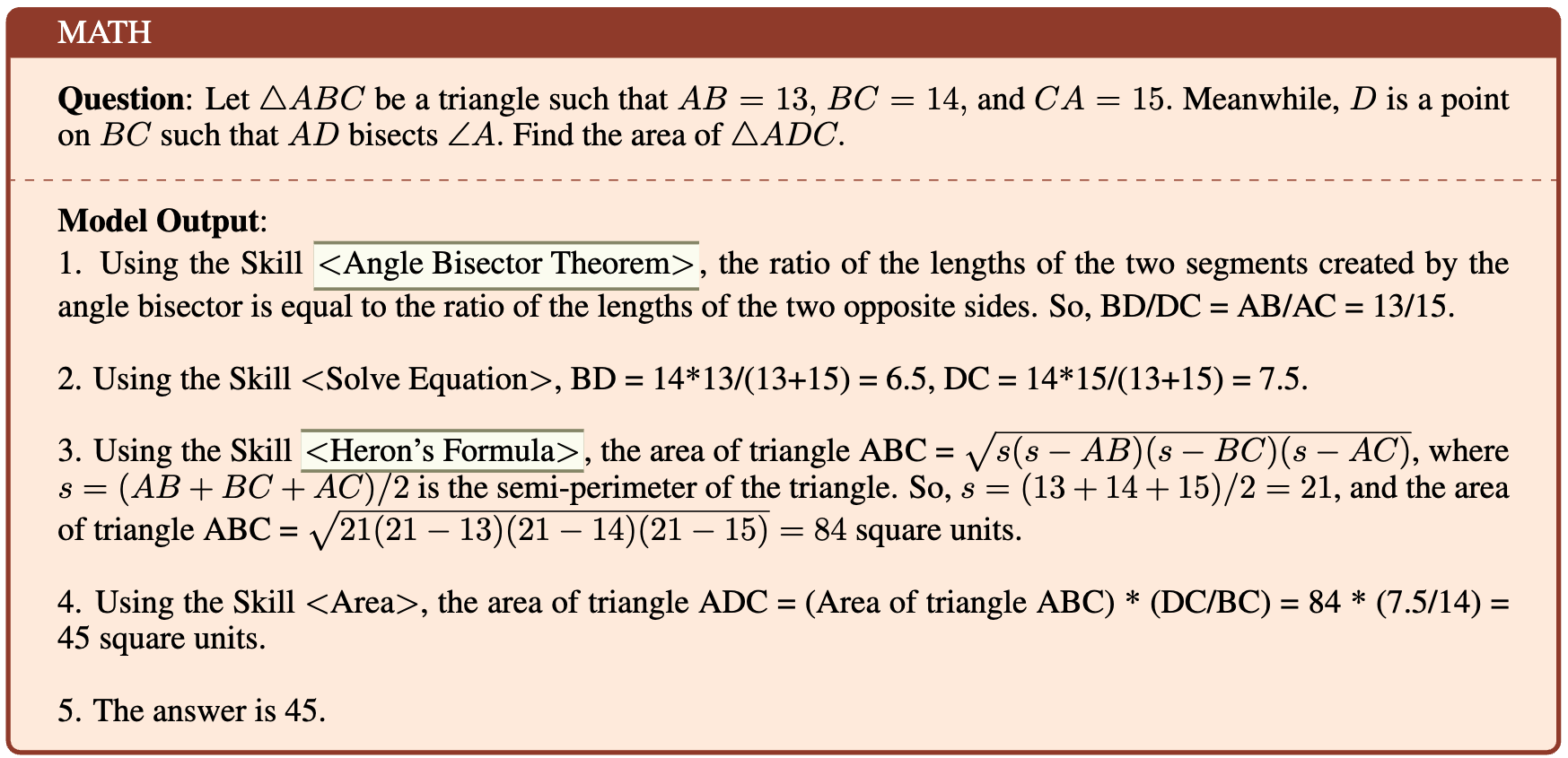}}
\caption{An example of the generated solution on the MATH task using SKiC. Intriguingly, the two highlighted skills $<$Angle Bisector Theorem$>$ and $<$Heron's Formula$>$ are neither provided in the SKiC context (see Figure \ref{Tab:math_skill}) nor used in any given exemplars. LLMs harness the internal skills in their pre-trained knowledge to solve the problem, where these two highlighted skill names are also generated automatically by the LLM.} \label{Tab:example_compose_math_skill_5}
\end{center}
\vskip -0.2in
\end{figure*}

\paragraph{Comparison to existing approaches}
Different from Chain-of-Thoughts, our SKiC provides explicit grounding on the foundational skills at each of the reasoning steps and also provides the relevant skills within the same context. Compared to recent prompting methods for handling compositional problems such as Least-to-Most (LtM) \citep{zhou2022least} and Decomp \citep{khot2022decomposed}, our SKiC is superior in several dimension: (i) Our SKiC is more general to solve extended sets of problems. Previous decomposing-based approaches like LtM and Decomp usually solve complex problems in a two-stage fashion by first decomposing the problem into a linear sequence of subproblems and then solving them sequentially. However, many of the tasks that have complex computation graphs such as multiplication and dynamic programming problems~\citep{dziri2023faith} cannot be decomposed in a simple manner, which makes these decomposition-based approaches less applicable. (ii) The decomposition operation can also be viewed as one basic skill in SKiC (see Figure~\ref{Tab:qa_skill} for an example in a question-answer task). (iii) SKiC solves the complex problems in a single stage, which could alleviate the error propagation compared to decomposition-based approaches that require multiple distinct stages. Due to the one-stage nature, our SKiC can replace other one-stage strategies such as the CoT in a plug-and-play manner. And it can be easily combined with other ensemble techniques such as self-consistency \citep{cot_wei_sc} and Progressive-Hint \citep{zheng2023progressive} to further boost the performance. Please refer to Appendix~\ref{sec:tool} for the relations to tool-using.

\paragraph{Construction of the skills} \label{Sec:skic_construction}
One important component in the above SKiC structure is the foundational skills. Note that these skills are not meant to be an exclusive coverage over all the necessary skills. Instead, they are intended to be used together with the compositional exemplars to demonstrate how to perform explicit and grounded composition. For this reason, we only need a limited number of in-context skills since they only need to be used together with a few (typically $2 \sim 10$) compositional exemplars. Therefore, the human effort involved in constructing these skills are generally minimal or at most comparable to other few-shot prompting approaches. Indeed, our experimental analysis shows that SKiC requires less number of demonstration examples. Morever, these skills can also be constructed automatically by prompting LLMs while still achieving good performance (see the results in Section \ref{sec:experiment:synergy} and more details in Appendix \ref{sec:more_details_construction}).

\paragraph{Grounding the composition}
As shown in Figure \ref{fig:example}, we explicitly ground the reasoning steps onto the corresponding skills in the compositional exemplars. Besides the in-context skills, we may also ground the reasoning steps to the internal skills not presented in the context, where the existence of these internal skills can be verified by prompting the LLMs with the skill information (see Appendix \ref{sec:more_details_construction}). Intriguingly, with SKiC, the LLMs can more actively tap into the vast reservoir of the internal skills they acquired during the pre-training stage in complex reasoning. In Figure \ref{Tab:example_compose_math_skill_5}, we demonstrate an example of the generated solution on the MATH task using SKiC. The two highlighted skills $<$Angle Bisector Theorem$>$ and $<$Heron's Formula$>$ are neither provided in the SKiC context (see Figure \ref{Tab:math_skill}) nor used in any given exemplars. LLMs automatically ground onto the (pre-trained) internal skills and compose them in their output reasoning steps. Notably, these two highlighted skill names are also automatically generated by the LLM.

\begin{figure*}[t]
\begin{center}
\centerline{\includegraphics[width=2.0\columnwidth]{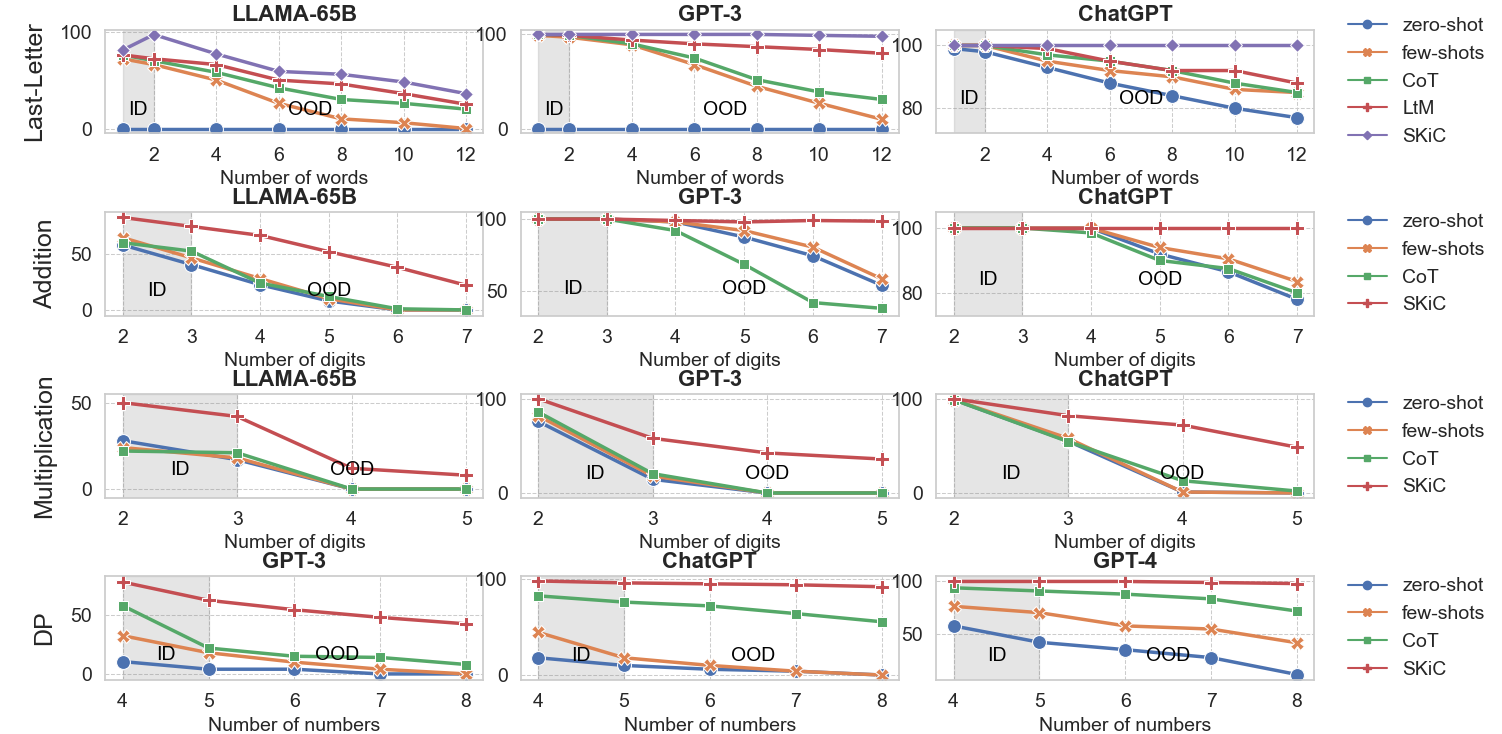}}
\caption{Accuracy on last letter concatenation, addition, multiplication, and dynamic programming. The gray area is in-distribution evaluation where the testing examples are with the same level of complexity as examples in the context, while the white area is out-of-distribution evaluation where the test set are increasingly harder problems.} 
\label{fig:main_results}
\end{center}
\vskip -0.2in
\end{figure*}

\begin{figure}[t]
\begin{center}
\centerline{\includegraphics[width=1.0\columnwidth]{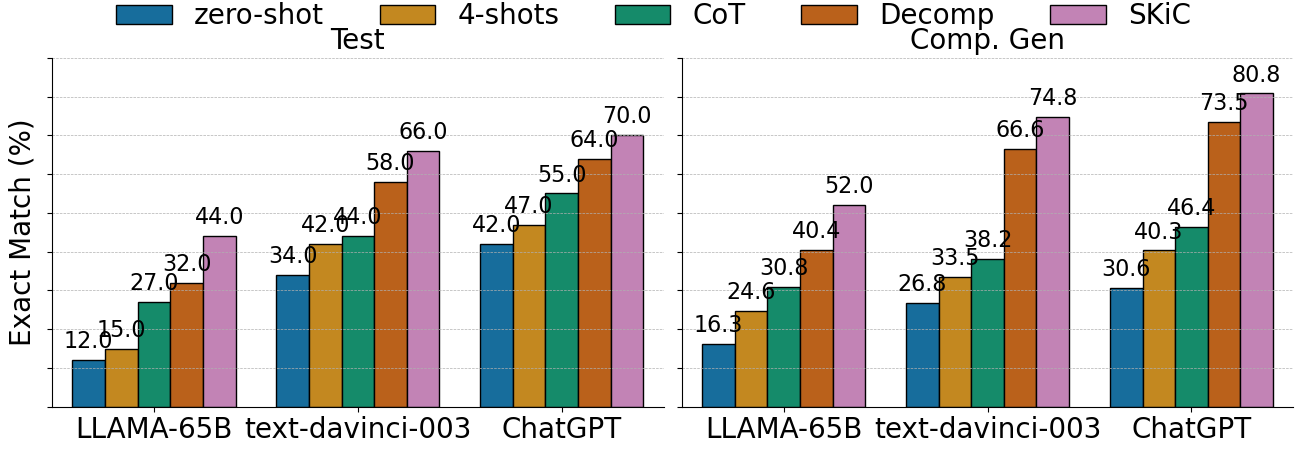}}
\caption{Exact Match on Commaqa-E. The ``Comp. Gen'' reports the results on the compositional questions.} 
\label{fig:QA_results}
\end{center}
\vskip -0.2in
\end{figure}

\section{Analysis of Compositional Abilities} \label{sec: experiments}
We perform experiments in two settings, where more details can be found in Appendix~\ref{Sec:setup}:

\textbf{Systematic Generalization: Composition over in-context skills},  where all the needed skills are provided in the context. We evaluate (i) last letter concatenation \citep{wei2022chain,zhou2022least,khot2022decomposed}, where the LLM needs to generate the concatenation of the last letter from a given list of words, (ii) addition and multiplication \citep{dziri2023faith}, where the LLM needs to generate the sum and product of two numbers, (iii) CommaQA-E \citep{khot2022decomposed}, where models need to answer multi-hop questions, and (iv) dynamic programming \citep{dziri2023faith}, where LLMs need to find the highest sum for a subsequence where no two numbers are adjacent. These tasks require only a limited skill set and we construct SKiC prompts manually in Figures~\ref{Tab:last_letter_skill}-\ref{Tab:compose_dp_skill}, with similar human effort as in CoT prompting.

\textbf{Complex Reasoning: Generalization beyond in-context skills},  where models need to harness skills beyond the context and tap into the internal skills for math reasoning like GSM8K~\citep{cobbe2021training} and MATH~\citep{hendrycks2021measuring}. 
For GSM8K, which are simpler problems that could be solved by basic math operations, we construct SKiC via human in Figures~\ref{Tab:gsm8k_skill}-\ref{Tab:compose_gsm8k_skill}. For MATH, which is a more challenging benchmark, we prompt the LLMs to generate the skills and then handcraft a few examples in Figures~\ref{Tab:math_skill},\ref{Tab:compose_math_skill} (see the second approach in Appendix~\ref{sec:more_details_construction}). The handcrafting effort involved here is comparable to other few-shot prompting approaches such as CoT.

We mainly compare SKiC with zero/few-shot standard prompting \citep{brown2020language}, CoT~\citep{wei2022chain}, Least-to-Most (LtM) \citep{zhou2022least}, and Decomp \citep{khot2022decomposed} on different LLMs including LLAMA \citep{touvron2023llama1}, GPT3 (text-davinvi-003) \citep{brown2020language}, ChatGPT and GPT4~\citep{openai2023gpt4}. For tasks in the second setting, we further compare our methods with Scratchpad \citep{nye2021show}, Learning-to-Program (LtP) \citep{guo2023learning},  ComplexCoT \citep{fu2022complexity} and ensemble strategies such as majority voting (maj1@k) \citep{lewkowycz2022solving}, Self-Consistency (SC) \citep{cot_wei_sc}, Progressive-Hint Prompting (PHP) \citep{zheng2023progressive}, and Code-based-Verification (CSV)\citep{zhou2023solving}. Note that all the exemplars in SKiC are either a subset of or the same as what have been used in baselines. 

\subsection{Near-Perfect Systematic Generalization}
We report the main results for last letter concatenation, addition \& multiplication, Commaqa-E and DP in Figures~\ref{fig:main_results}-\ref{fig:QA_results}. Additional results can be found in Appendix~\ref{sec:detailed_results}. Standard zero/few-shot prompting generalizes poorly on the problems that are harder than the exemplars in the prompting context. 
CoT, LtM and Decomp improve the overall performance but still degrade quickly over harder inputs. SKiC significantly boosts the performance in harder cases. 
Notably, SKiC achieves nearly perfect generalization on tasks like last letter concatenation, addition, and dynamic programming with text-davinci-003, ChatGPT or GPT4. 
These significant improvements highlight the importance of in-context skills and explicit grounding in eliciting compositionality.
Examples of the generated answers with SKiC can be found in Figures~\ref{Tab:example_last_letter_skill}-\ref{Tab:example_compose_dp_skill}.


\begin{figure}[t]
\begin{center}
\centerline{\includegraphics[width=1.0\columnwidth]{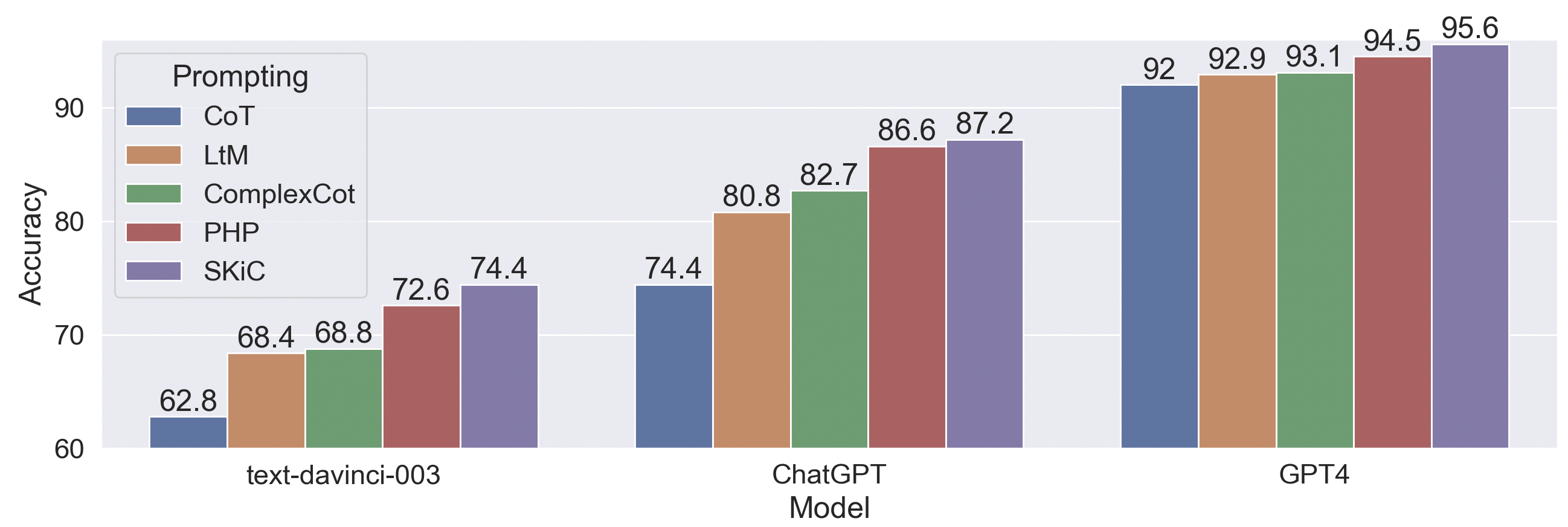}}
\caption{The accuracy on GSM8K tasks.} 
\label{fig:gsm8k_results}
\end{center}
\vskip -0.2in
\end{figure}

\definecolor{red5}{rgb}{0.55,0.95,0.55}

\begin{table*}[t]
\caption{Accuracy and internal skill activation rate on the MATH. 
} \label{Tab:math_results}
\centering
\scalebox{0.6}{%
\begin{tabular}{c|c|c|ccccccc|c} \toprule
\textbf{Model}        & \textbf{Prompting }      & \textbf{Ensemble} & \textbf{Pre-Algebra}   & \textbf{Geometry}      & \textbf{Inter-Algebra} & \textbf{Algebra}       & \textbf{Probability}   & \textbf{Pre-Calculus}  & \textbf{NumTheory}     & \textbf{Overall}       \\  \midrule  \midrule

PaLM-2  & CoT             & SC       & -             & -             & -             & -             & -             & -             & -             & 48.8          \\ 
Minerva-540B & CoT, Scratchpad & maj1@k    & 71.1          & 42.0          & 27.1          & 72.7          & 43.5          & 34.5          & 36.3          & 50.3          \\ 
ChatGPT      &  Verification     & CSV      & 58.9          & 22.0          & 14.8          & 45.6          & 35.2          & 13.0          & 33.5          & 34.7         \\  
GPT-4        & Verification    & CSV      & 76.2          & 38.6          & 25.3          & 70.4          & 57.0          & 28.6          &  53.5         & 51.8          \\  
ChatGPT      & ComplexCoT      & PHP      & 57.7          & 25.4          & 17.1          & 49.1          & 33.7          & 16.1          & 35.1          & 36.5          \\  
GPT-4        & ComplexCoT      & PHP      & 73.8          & 41.9          & 26.3          & 73.4          & 56.3          & 29.8          & 55.7          & 53.9          \\  \midrule \midrule

PaLM-2        & CoT             & \XSolidBrush     & -             & -             & -             & -             & -             & -             & -             & 34.3          \\
Minerva-540B & CoT, Scratchpad & \XSolidBrush     & 54.9          & 26.7          & 13.6          & 51.2          & 27.9          & 18.0          & 21.2          & 33.6          \\ \midrule

\multirow{4}{*}{ChatGPT}        & CoT, LtP      & \XSolidBrush     & 52.3          & 22.5          & 16.9          & 49.6          & 30.2          & 16.3          & 29.8         & 31.1          \\
& ComplexCoT      & \XSolidBrush     & 53.8          & 22.3          & 14.6          & 49.1          & 29.7          & 16.8          & 33.4          & 34.1  \\
      & SKiC (\textbf{Ours})           & \XSolidBrush     & 62.0 \small{\colorbox{red5}{$\uparrow8.2$}}           & 30.1 \small{\colorbox{red5}{$\uparrow7.8$}}          & 17.8 \small{\colorbox{red5}{$\uparrow3.2$}}        & 57.9  \small{\colorbox{red5}{$\uparrow8.8$}}        & 38.2 \small{\colorbox{red5}{$\uparrow8.5$}}         & 23.0 \small{\colorbox{red5}{$\uparrow6.2$}}         & 35.5  \small{\colorbox{red5}{$\uparrow2.1$}}        & 40.6 \small{\colorbox{red5}{$\uparrow6.5$}}         \\ \cmidrule{2-11}
        
        & \multicolumn{2}{c|}{\textit{Internal Skill Activation Rate}}    & \textit{6.5} & \textit{19.0} & \textit{13.2} & \textit{5.7} & \textit{9.1} & \textit{45.2} & \textit{7.8} & \textit{14.9} \\ 
       \midrule

\multirow{4}{*}{GPT4}        & CoT             & \XSolidBrush     & -             & -             & -             & -             & -             & -             & -             & 42.2          \\
        & ComplexCoT      & \XSolidBrush     & 71.6          & 36.5          & 23.4          & 70.8          & 53.1          & 26.7          & 49.6          & 50.3          \\
        & SKiC (\textbf{Ours})           & \XSolidBrush     & \textbf{79.7} \small{\colorbox{red5}{$\uparrow8.1$}}  & \textbf{43.6} \small{\colorbox{red5}{$\uparrow7.1$}}  & \textbf{29.5} \small{\colorbox{red5}{$\uparrow6.1$}}  & \textbf{74.6} \small{\colorbox{red5}{$\uparrow3.8$}}  & \textbf{58.2} \small{\colorbox{red5}{$\uparrow5.1$}}  & \textbf{36.6} \small{\colorbox{red5}{$\uparrow9.9$}}  & \textbf{55.9} \small{\colorbox{red5}{$\uparrow6.3$}}  & \textbf{56.4} \small{\colorbox{red5}{$\uparrow6.1$}}  \\  \cmidrule{2-11}
        
        & \multicolumn{2}{c|}{\textit{Internal Skill Activation Rate}}    & \textit{12.7} & \textit{37.0} & \textit{33.4} & \textit{16.0} & \textit{4.4} & \textit{65.5} & \textit{12.1} & \textit{24.3} \\ 
        \bottomrule     
\end{tabular}
}
\end{table*}

\subsection{Enhanced Complex Reasoning}
Figure~\ref{fig:gsm8k_results} shows the significantly boosted accuracy on GSM8K by SKiC compared to other baselines, even with incomplete skills in  SKiC prompts. We observe several important generalization behaviors: (i) generated reasoning steps effectively utilize the provided skills that are not demonstrated in the compositional examples (Figure~\ref{Tab:example_compose_gsm8k_skill_1}), (ii) generated reasoning steps successfully employ skills that are not included in the prompts but may exist within the pre-trained knowledge of the LLM (Figures~\ref{Tab:example_compose_gsm8k_skill_2}-\ref{Tab:example_compose_gsm8k_skill_3}). They suggest that, with SKiC, LLMs can be taught to use the skills provided in the context as well as from their pretrained knowledge to solve math problems via compositionality. 

Accuracy on MATH is reported in Table~\ref{Tab:math_results}. With SKiC constructed in a semi-automated manner, models could explicitly ground the reasoning steps to both in-context skills and their internal knowledge to resolve math problems, leading to SKiC's superior performances. 
We also show the \emph{internal skill activation rate} that measures the percentage of skills utilized in the generated reasoning steps that originates from pre-trained knowledge (rather than being introduced in the SKiC prompt). It further verifies that SKiC allows the LLMs to generalize beyond the in-context skills and more actively invoke the massive reservoir of internal capabilities in LLMs (e.g., 24\% of skills utilized in the output reasoning steps are from the GPT4 internal knowledge) --- see Figures~\ref{Tab:example_compose_math_skill_1}-\ref{Tab:example_compose_math_skill_4} for more examples, where the reasoning process carried out by the LLM effectively utilize both in-context and internal skills. The frequently used in-context and internal skills are illustrated in Table~\ref{Tab:math_top_skills} in Appendix.

\subsection{Synergy between Skills and Composition}
\label{sec:experiment:synergy}

\begin{table}[t]
 \caption{Accuracy on RTE and Last Letter  (12 words) with ChatGPT models using skills crafted by human or skills discovered by LLMs in SKiC.} \label{Tab:discover_main}
\centering
\begin{tabular}{c|c|c} \toprule
\textbf{Methods}                & \textbf{RTE} & \textbf{Last Letter}  \\  \midrule \midrule
COT & 85.2 &72.5 \\ \midrule
SKiC by Human &-  & \textbf{100.0}  \\
SKiC by LLM & \textbf{89.8}  & \textbf{100.0} \\   \bottomrule    
\end{tabular}
\end{table}

\paragraph{Skills from Human vs. Skills Discovered by Models}
We conduct experiments to show that the skills can be discovered automatically by LLMs, which makes our SKiC more applicable to a wider range of tasks. We provide ChatGPT with examples from the training sets of RTE~\citep{wang2018glue} and last letter tasks, and instruct it to discover the skills from the examples to solve the tasks, which results in skills such as \textit{Context Understanding} and \textit{Inference Evaluation} for RTE, and  \textit{Identify Words}, \textit{Determine Last Letters}, \textit{Concatenate Last Letters}, \textit{Form New Sequence} for last letter. Based on the summarized skills from LLMs, we then construct SKiC prompts. The results are shown in Table~\ref{Tab:discover_main}, which demonstrates the effectiveness of SKiC with automatically discovered skills.

\begin{table}[t]
\caption{Accuracy and internal skill activation rate on MATH with two variants of SKiC on ChatGPT: the skills are generated from (i) ChatGPT and (ii) GPT-4. 
} \label{Tab:math_simple_ablation_results}
\centering
\begin{tabular}{c|c|c} \toprule
    \textbf{Metric}   &\textbf{Source of Skills}   & \textbf{Overall}        \\  \midrule  \midrule

\multirow{2}{*}{Accuracy} &GPT4             & 38.9          \\  

& ChatGPT   & \textbf{40.6} \\ \midrule 

\multirow{2}{*}{\begin{tabular}[c]{@{}c@{}}Internal Skill \\ Activation Rate\end{tabular}} &GPT4         & 12.5   \\   
        
    &ChatGPT              & \textbf{14.9} \\ \bottomrule
\end{tabular}

\end{table}

\paragraph{Skills from Stronger Model vs. Skills from the Same Generative Model}
Another important question we want to understand is whether it is beneficial to generate the in-context skills from the same foundation model used for prediction. We prompt the ChatGPT using the SKiC constructed from itself or the stronger GPT-4 (i.e., the in-context skills are generated by GPT-4). The accuracy and the internal skill activation rate on MATH are reported in Table~\ref{Tab:math_simple_ablation_results} (see Table~\ref{Tab:math_ablation_results} for the complete result).
With the skills prompted from itself, we observe improved accuracy and skill activation rate. This suggests that (i) aligning the model that is used to prompt the in-context skills and the model that is used to generate answers helps the models' capability to link and utilize internal skills, and (ii) activating more internal skills leads to higher performance for complex problems.

\begin{table}[t]
 \caption{Accuracy of MATH and FOLIO when using prompts designed for GSM8K with ChatGPT models.  } \label{Tab:different_tasks_main}
\centering
\begin{tabular}{c|c|c} \toprule
\textbf{TASK}                & \textbf{COT for GSM8K} &\textbf{SKiC for GSM8K}     \\  \midrule \midrule
MATH    & 28.2 &\textbf{31.34} \\
FOLIO   & 68.8 &\textbf{72.5} \\   \bottomrule    
\end{tabular}
\end{table}

\paragraph{Generalization to New Tasks}
We further show that SKiC generalizes better than CoT when we apply a prompt (originally designed for a different task) directly to new unseen tasks. To see this, we apply the prompts designed for GSM8K to MATH (competition-level math reasoning) and to FOLIO (logical inference) \citep{han2022folio}, which are unseen new tasks (see Table~\ref{Tab:different_tasks_main}). Compared to CoT, SKiC shows better cross-task transfer abilities.

\begin{table}[t]
 \caption{Accuracy on DP (8 numbers) of SKiC with ChatGPT after removing different components.} \label{Tab:different_components_main}
\centering
\begin{tabular}{c|c} \toprule
\textbf{Methods}                & \textbf{Dynamic Programming}   \\  \midrule \midrule
COT    & 72.0  \\ \midrule
SKiC    & \textbf{98.0}  \\ 
- skill   & 94.0  \\
- skill grounding    & 82.0  \\ \bottomrule    
\end{tabular}
\end{table}

\paragraph{Ablation Analysis of SKiC Components}
In our work,  we discover that besides step-by-step reasoning, explicit grounding is another key factor to elicit compositional generaization, demonstrated by significantly better performances of SKiC. We perform ablation study to highlight the finding (the importance of skills and skill grounding). We compare SKiC with the settings where (i) we remove the skills but keep the skill grounding in reasoning steps and (ii) we remove the skill grounding in reasoning steps but keep the basic skill introduction in the front. The performance on Dynamic Programming is shown in Table~\ref{Tab:different_components_main}. Removing either parts would lead to performance drop, which further indicates the importance of both skills and skill grounding to for compositional generalization.

\begin{table}[t]
 \caption{Accuracy of different sets of few-shot exemplars in CoT and SKiC on the last letter with ChatGPT.  } \label{Tab:different_prompts_main}
\centering
\small
\begin{tabular}{c|c|c} \toprule
\textbf{Examples in Prompts}                & \textbf{COT} &\textbf{SKiC}     \\  \midrule \midrule
'apple, banana'; 'apple, pie'    & 91.4 &\textbf{100.0} \\
 'math, code'; 'science, computer'   & 92.5 &\textbf{100.0} \\
 'ashc, edhoh'; 'shbod, wojois'     & 90.8 &\textbf{100.0} 
 \\   \bottomrule    
\end{tabular}
\end{table}

\begin{table}[t]
 \caption{Accuracy of different orders of few-shot exemplars in CoT and SKiC on GSM8K with ChatGPT.  } \label{Tab:different_orders_main}
\centering
\begin{tabular}{c|c|c} \toprule
\textbf{Order of Examples}                & \textbf{COT} &\textbf{SKiC}     \\  \midrule \midrule
Random order 1   & 74.4 &\textbf{87.2} \\
Random order 2   & 73.8 &\textbf{86.9} \\
Random order 3    & 73.0 &\textbf{87.8} 
 \\   \bottomrule    
\end{tabular}
\end{table}

\paragraph{Robustness to Few-shot Exemplars}
We evaluate the robustness of SKiC to the choices and the orders of exemplars in Tables~\ref{Tab:different_prompts_main}-\ref{Tab:different_orders_main}, respectively, where SKiC is robust against the selection of few-shot exemplars and shows a similar level of robustness as CoT while achieving better overall performance.

\begin{figure*}[t]
    \centering
    \includegraphics[width=2.0\columnwidth]{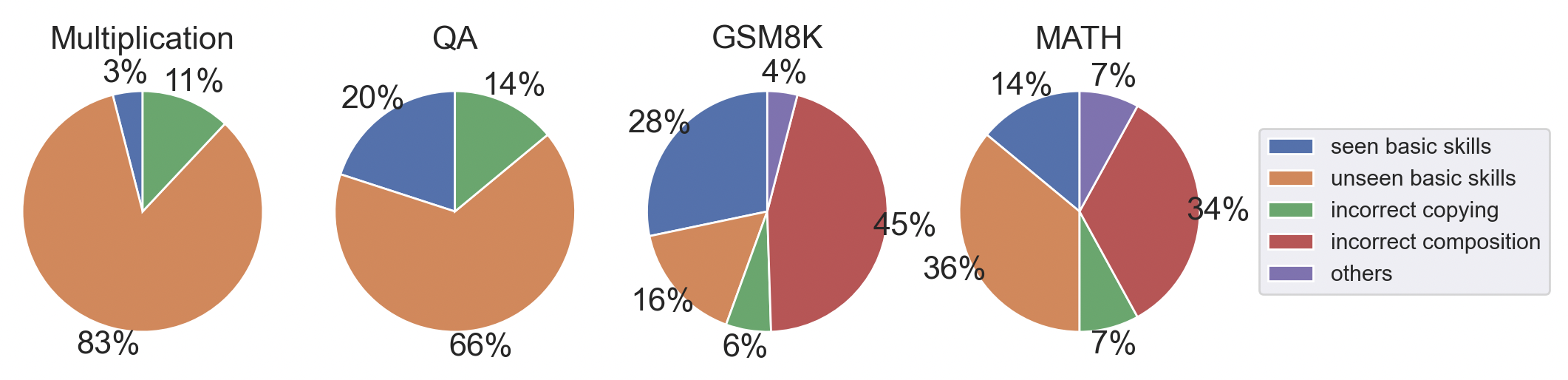}
    \caption{Error distributions in Multiplication, QA, GSM8K and MATH tasks.}
    \label{fig:errors}
\end{figure*}

\subsection{Error Analysis}
We perform error analysis on the tasks that are still far away from (nearly) perfect generalization when applying SKiC on ChatGPT --- multiplication, question answering, GSM8K and MATH. For each task, we randomly sample 50 error cases and perform an examination of them. We summarize five types of errors: (i) seen basic skills: errors arise due to a lack of mastery of the skills in context, (ii) unseen basic skills: errors caused by the absence of skills in context, particularly when these skills do not exist in the pre-trained knowledge, (iii) incorrect composition: errors of incorrect composition or reasoning over the skills, (iv) incorrect copying: copying or merging errors between different steps, (v) others: such as incorrect labels in the test set. 

The distributions are visualized in Figure~\ref{fig:errors}. We observe that (i) the most common errors arise from unseen basic skills,
(ii) a lack of mastery of the basic skills leads to more errors when there are more complex or more basic skills to be used (for example, the question decomposition capability in the CommaQA-E task is generally a complex skill, and the GSM8K and MATH dataset requires more basic skills), 
(iii) incorrect composition is a major error type for tasks that require more complex reasoning steps such as GSM8K, 
(iv) copying errors become more prevalent when there are more reasoning steps with longer context, and (v) math reasoning generally requires a wider variety of skill compositions, and the way of composition varies significantly from one problem to another, making it considerably harder to master the appropriate skill composition for each problem. 

\begin{figure}[t]
\begin{center}
\centerline{\includegraphics[width=1.0\columnwidth]{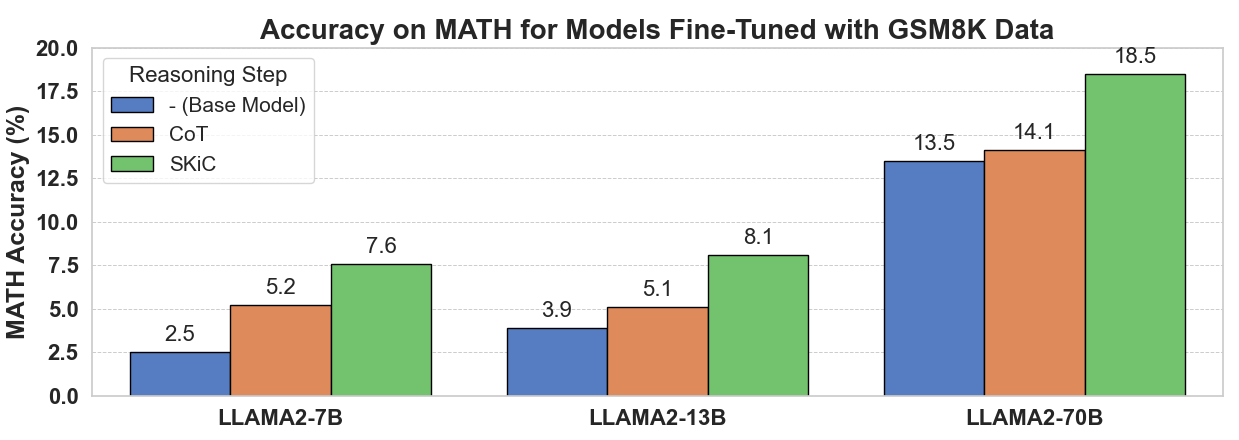}}
\caption{Generalization from GSM8K to MATH.} 
\label{fig:finetun_results}
\end{center}
\vskip -0.2in
\end{figure}

\section{Beyond In-Context Learning}
\label{sec:ift}

Inspired by the above in-context learning study, we show that instruction tuning data constructed with SKiC structure can be utilized to fine-tune LLMs, enhancing their easy-to-hard generalization capabilities. Specifically, we generate training data by using GPT4 to produce answers for GSM8K problems with SKiC prompts. This ensures that the reasoning steps for each GSM8K problem are explicitly grounded on basic skills, as illustrated in Figure~\ref{Tab:example_compose_gsm8k_skill_2}-\ref{Tab:example_compose_gsm8k_skill_3}. Using the GSM8K data annotated with SKiC-structure reasoning steps, we fine-tune Llama2 models and evaluate their performance on MATH dataset, which consists of significantly more challenging evaluation problems compared to the training problems from GSM8K. The results, shown in Figure~\ref{fig:finetun_results}, indicate that fine-tuning with SKiC data significantly improves accuracy on MATH compared to training data annotated with CoT reasoning steps (also by GPT4). This demonstrates that models fine-tuned with SKiC reasoning steps achieve better generalization to complex and challenging test cases. These findings suggest that SKiC could potentially replace CoT in instruction tuning, eliciting stronger reasoning capabilities and enabling better weak-to-strong generalization.

\section{Related Work}
\label{sec: related works}
There has been a long history of studies on compositional generalization \citep{lake2018generalization,jia2016data, andreas2019good,lake2018generalization,ouyang2023compositional,keysers2020measuring,chen2020compositional,dziri2023faith,shao2023compositional,saparov2022language,nye2021show,welleck2022naturalprover,dong2019neural,schwarzschild2021can}. Different types of approaches have been developed to solve compositional generalization. One widely studied approach is neuro-symbolic methods \citep{dong2019neural,schwarzschild2021can}, which blend symbolic and distributed representations for modeling the reasoning process. A recent line of work that has gained significant traction is to prompt large language models to unlock its potential compositional generalization abilities \citep{nye2021show,zhou2022least,khot2022decomposed,dua2022successive,dziri2023faith}. For example, least-to-most prompting \citep{zhou2022least} and decomposed prompting \citep{khot2022decomposed} boosts compositional generalization by first decomposing a difficult problem into a sequence of easy-to-hard problems and then solving them sequentially. 
However, the performance still degrade quickly over increasingly harder problems. Moreover, their applications are limited to a class of problems that can be decomposed into a set of subproblems. For more general complex problems, where the subproblems are highly nested (e.g., the ones shown in \citet{dziri2023faith}), it becomes quite challenging to construct the prompts and the exemplars. Recent work \citep{zhang2023cumulative,zhou2023solving} have also explored multiple agents for solving complex problems. Unlike these multi-stage/agents prompting methods, which require multiple calls of multiple LLM in inference process, our proposed Skills-in-Context prompting is a simple one-stage/single-agent strategy that can be used in a plug-and-play manner to replace existing standard or CoT prompting. While concurrent work \citep{zhou2024self,zheng2023take} also highlights the appearance of skills in prompts, our studies further show the importance of explicit grounding to basic skills in reasoning steps.

\section{Conclusion}
\label{sec: conclusion}
In this work, we examine how to elicit compositional generalization abilities in LLMs. Specifically, within the in-context learning framework, we find that it is crucial to explicitly ground each of the reasoning steps on the foundational skills. To facilitate this, it is important to demonstrate both the foundational skills and the compositional examples grounded in these skills within the same prompt context. We refer to this prompt structure as skills-in-context (SKiC). SKiC demonstrates strong (near-perfect) systematic generalization abilities across many tasks and enhanced complex reasoning capabilities. Notably, with SKiC, the LLMs could generalize beyond the skills provided in the prompting context and learns to activate the skills and knowledge that are acquired through earlier pre-training stages for solving unseen complex problems. Furthermore, SKiC structure could be utilized in fine-tuning to improve the easy-to-hard generalization. 

\section{Limitations}
In this work, we follow the previous work \citep{dziri2023faith, zhou2022least} and mainly focus on the compositional (easy-to-hard) generalization. Specifically, the in-distribution/seen tasks here means the testing samples are sampled from the same problem size \citep{dziri2023faith}. For example, we demonstrate examples of 2-digit addition, and then test it over unseen samples that are also from 2-digit addition. In contrast, the out-of-distribution/unseen tasks here are defined to be the harder unseen variants of the problem.  For example, the testing samples of 5-digit additions are the harder variant of the problem that are not seen in the context examples. And we utilize the SKiC to improve such easy-to-hard compositional generalization and complex reasoning tasks compared to previous methods. In the era of LLMs, although it is challenging to investigate whether the LLMs have been pre-trained on some of the tasks, we believe that even if some of the tasks could be crawled into the pretraining corpus, they are mostly general and simple examples (e.g., last letters of 4 or 5 words)  rather than the harder cases that we tested on (e.g., last letters of 12 words). This is also demonstrated in the zero-shot performances on the harder cases: for example, the zero-shot performances of ChatGPT on last-letter, addition, multiplication and dynamic programming are quite low (lower than 50\% in most of the cases)). With our SKiC, the easy-to-hard generalization capability is significantly boosted to even near-perfect generalization, while other strong prompting methods such CoT and Least-to-Most cannot do so.

Furthermore, despite the promising results demonstrated by Skills-in-Context (SKiC), there are several limitations and challenges to explore in future work. First, from our error analysis, there are several key directions for further improvements: (i) providing high-quality basic skills and illustrations to improve the execution quality of these basic skills, (ii) expanding the range of task-related basic skills to prevent errors caused by unseen skill, (iii) providing more examples of how to compose basic skills, especially for more complex tasks, and (iv) utilizing better foundation models that can handle longer context and have a more extensive set of well-mastered skills in their pre-pretrained knowledge. Second, while SKiC has shown strong performance in problems with relatively clear and limited skill sets, scaling it to more complex domains where the number and variety of required skills are vast remains challenging. The manual or semi-automatic approach to skill distillation may not be feasible for problems requiring a broad and intricate combination of skills, such as those in dynamic, real-world scenarios. Future work could explore how to improve the adaptation through fine-tuning with SKiC structures. Third, our approach focuses primarily on utilizing internal skills without extensive reliance on external tools or resources. While this reduces inference latency and leverages the internal knowledge of LLMs, it may limit the applicability of SKiC in scenarios where external tools could provide significant advantages, such as in real-time data retrieval or complex calculations that exceed the capabilities of the model's internal knowledge base. Future work could also utilize external tools to further improve the performance.

\bibliography{custom}

\clearpage
\appendix

\section{Comparison to Existing In-Context Learning Strategies}
\label{appendix:comparison}
\begin{figure*}[ht]
\vskip 0.2in
\begin{center}
\centerline{\includegraphics[width=1.9\columnwidth]{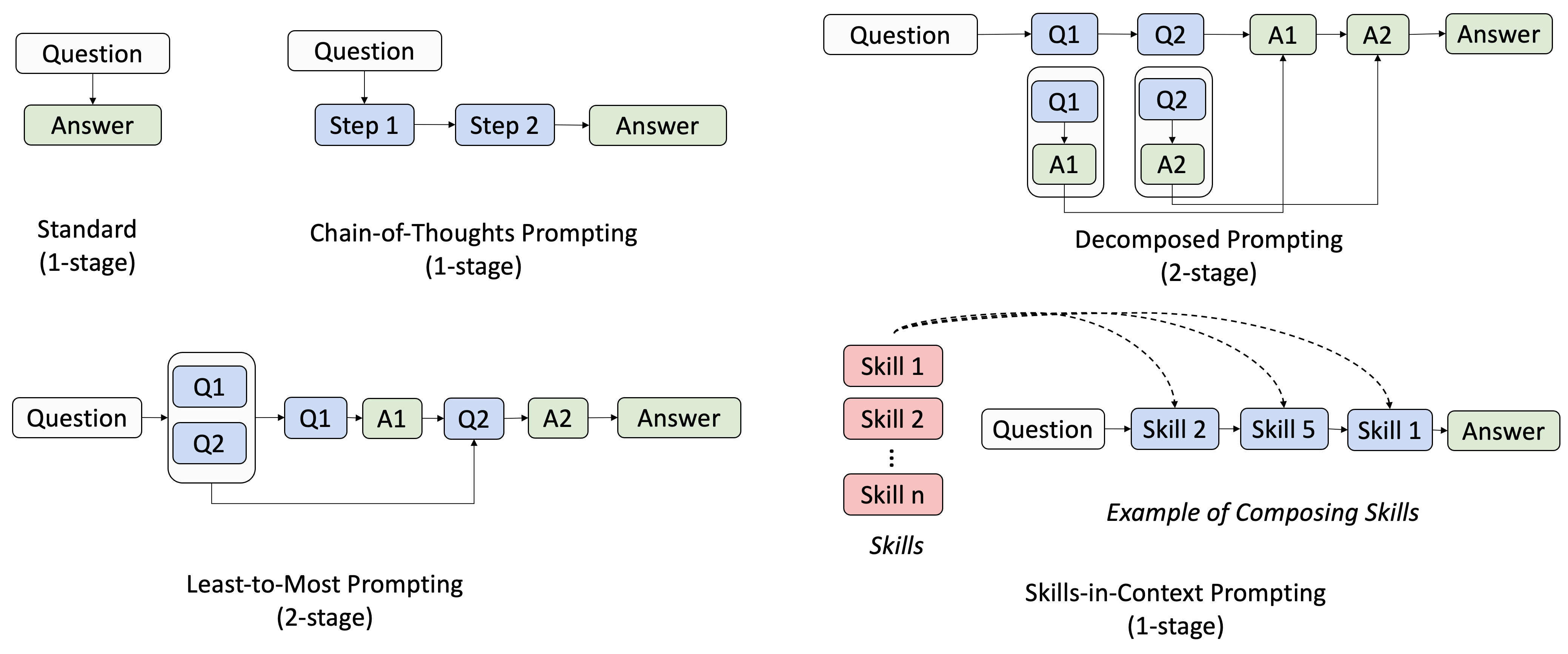}}
\caption{The building blocks of different prompting strategies. Blue cells stand for different intermediate steps, green cells denote the answers to the asked question, and red cells refer to the provided skills in our Skills-in-Context prompting. A block of several cells represents one distinct stage in a two-stage prompting strategy (e.g., problem decomposition stage in the Least-to-Most prompting). Standard prompting provides only labeled exemplars in the context. Chain-of-Thoughts prompting further provides a step-by-step rationale preceding the answer. Decomposed prompting is a two-stage prompting method, which first breaks the questions into sub-problems, and then utilizes standard or Chain-of-Thoughts prompting to solve each sub-problem sequentially to derive the final answer. Least-to-Most prompting adopts a two-stage strategy: it first generates multiple questions in an easy-to-hard manner, and then sequentially answers each of them until solving the original question. In contrast, our Skills-in-Context prompting is a simple one-stage prompting, which places both the (basic) skills and the demonstrations of how to compose them into solutions within the same prompt context. This teaches the LLM how to explicitly and adeptly ground each reasoning step onto the skills (illustrated in dashed lines), which unleashes strong synergies between skills and composition capabilities in LLMs, leading to strong compositionality over unseen harder problems. 
} 
\label{fig:methods}
\end{center}
\vskip -0.2in
\end{figure*}

Figure~\ref{fig:methods} visualizes the differences between our proposed SKiC prompting and the previous related prompting methods. Different from Chain-of-Thoughts prompting, our SKiC prompting provides explicit grounding on the basic skills for reasoning steps towards final answers. Compared to recent prompting methods for handling compositional problems such as Least-to-Most prompting (LtM) \citep{zhou2022least} and Decomp \citep{khot2022decomposed}, our SKiC is superior in several dimensions: (i) Our SKiC prompting is more general to solve extended sets of problems. Previous decomposing-based approaches like LtM and Decomp usually solve complex problems in a two-stage fashion by first decomposing the problem into a linear sequence of subproblems and then solving them sequentially. However, many of the tasks that have complex computation graphs such as multiplication and dynamic programming problems~\citep{dziri2023faith} cannot be easily and fully decomposed in one stage, which makes it hard to apply these decomposition-based approaches. (ii) The decomposition operation can also be viewed as one basic skill in our SKiC prompt (for example, we view the decomposition operation as one of the skills in the question-answer task in Figure~\ref{Tab:qa_skill}). (iii) SKiC solves the complex problems in a single stage, which could alleviate the error propagation compared to decomposition-based approaches that require multiple distinct stages.

Due to the one-stage nature, our SKiC prompting can replace other one-stage strategies such as the CoT promptings in a plug-and-play manner. And it can also be easily combined with other ensemble techniques such as self-consistency \citep{cot_wei_sc} and Progressive-Hint Prompting \citep{zheng2023progressive} to further boost the performance.

\section{Details about the Construction of Skills} \label{sec:more_details_construction}

\begin{figure*}[t]
\begin{center}
\centerline{\includegraphics[width=1.9\columnwidth]{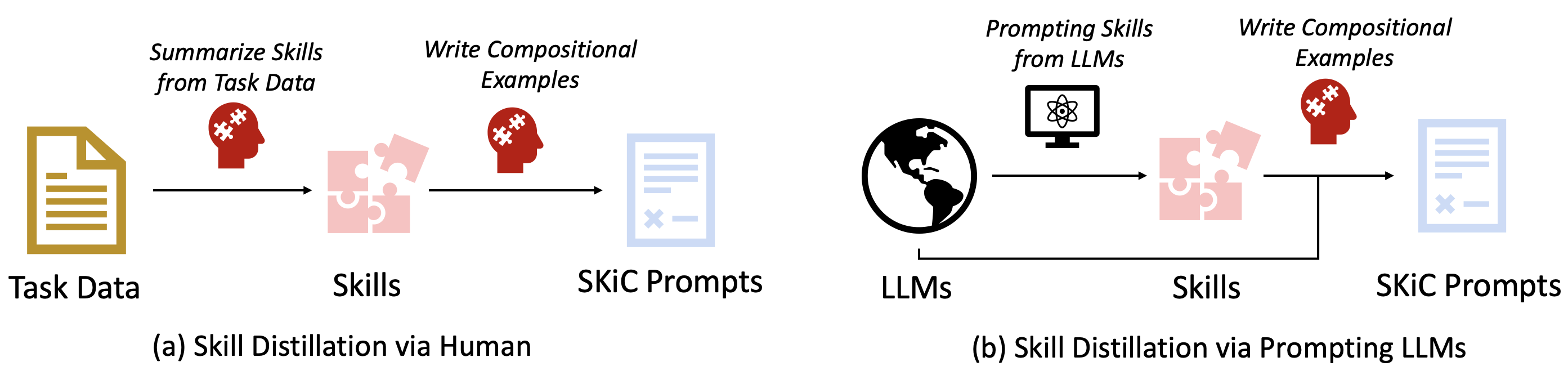}}
\caption{Two approaches to creating SKiC prompts, depending on how we distill the skills. (a) We \emph{manually} summarize the skills from the sample problems, and then construct the compositional exemplars on how to compose these skills. (b) We prompt the LLMs to \emph{automatically} generate the necessary skills, followed by human review. Then we craft the compositionl exemplars by grounding their reasoning steps onto either the provided in-context skills or the inherent skills within the LLMs. 
} 
\label{fig:construction}
\end{center}
\end{figure*}

One important step in constructing SKiC is to distill the (basic) skills that might be needed for solving problems associated with a task. We introduce two approaches (shown in Figure~\ref{fig:construction}):

\paragraph{Distill Skills via Human} 
Similar to previous prompting techniques, this is a fully manual approach, where the basic skills are manually summarized from a few (less than 10) problems. For example, given several samples from the last-letter-concatenation task, we manually identify that ``words\_to\_list'' and ``last\_letter'' are common skills to be used. Based on the discovered skills, we add a few ($1 \sim 2$) simple examples to illustrate these basic skills alone. Once the in-context skills are constructed, we add the compositional examples to demonstrate the composition of these skills to solve a problem (Figure~\ref{fig:example}). This approach puts all the essential skills in the context and is generally applicable to narrow domain problems that require the composition of limited skills for solving harder problems. It is also beneficial for semi-parametric LLMs, which can dynamically access the most relevant skills from external memories based on each input instance and integrate them into the problem context \citep{borgeaud2022improving,pan2023knowledgeincontext}.

\paragraph{Distill Skills via Prompting LLMs} 
More efficiently, we could automatically construct the basic skills by prompting the LLMs to directly generate the fundamental skills or summarize the necessary skills from given examples. For instance, when identifying the skills to address the MATH task \citep{hendrycks2021measuring}, we prompt LLMs with phrases like ``basic skills in Algebra''. This leads the model to generate basic skills such as ``Factoring'' (see Figure~\ref{Tab:math_skill}). Next, we construct the compositional examples by grounding the reasoning steps on the skills. It is worth noting that an exemplar might require skills not explicitly presented in the context. In these instances, we anchor the reasoning to inherent skills within the LLMs. For example, in the compositional exemplar showcased in Figure~\ref{Tab:compose_math_skill}, aside from leveraging in-context skills like``Sub'', it also employs skills like ``Pascal's Triangle'' --- a capability not present in the context but inherently known to the LLM. Such a construction of the exemplars will encourage the model to generalize beyond the in-context skills and compose solutions from the internal capabilities as well --- see Figure~\ref{Tab:example_compose_math_skill_5} for an example of the generated solution that activates the internal skills $<$Angle Bisector Theorem$>$ and $<$Heron's Formula$>$. To be more specific, for every problem in the MATH task, around 24\% of the skills, as shown in Table~\ref{Tab:math_results}, applied in the reasoning steps stem from the LLM's internal pre-trained knowledge (see Table~\ref{Tab:math_top_skills} for the most frequently used internal skills). The ability to harness both in-context skills and inherent capabilities is crucial for addressing complex reasoning problems, which typically require varied compositions across a broad spectrum of skills. Manually enumerating every required skill within a prompt context is often impractical. Meanwhile, LLMs have accumulated a vast reservoir of knowledge and skills during their pre-training. Leveraging these internal competencies can unlock significant potential, allowing LLMs to tackle even more complex challenges.

\begin{table*}[t]
 \caption{Accuracy on different evaluation subsets of the last-letter-concatenation task. The testing problems with 1 and 2 words are in-distribution evaluation, while the ones with $4\sim 12$, 50 and 100 words are (harder) out-of-distribution evaluations. } \label{Tab:last_letter_results}
\centering
\scalebox{0.8}{
\begin{tabular}{c|c|c|cc|ccccccc} \toprule
\textbf{Model}                & \textbf{Prompting} &\textbf{\#-shots}     & \multicolumn{1}{c}{\textbf{1}} & \multicolumn{1}{c|}{\textbf{2}} & \multicolumn{1}{c}{\textbf{4}} & \multicolumn{1}{c}{\textbf{6}} & \multicolumn{1}{c}{\textbf{8}} & \multicolumn{1}{c}{\textbf{10}} & \multicolumn{1}{c}{\textbf{12}} & \multicolumn{1}{c}{\textbf{50}} & \multicolumn{1}{c}{\textbf{100}} \\ \midrule \midrule
\multirow{5}{*}{LLAMA-65B} & zero-shot  &0        & 0                     & 0                     & 0                     & 0                     & 0                     & 0                      & 0  & - & -                    \\
                           & 4-shots   &4      & 72.0                    & 66.0                    & 50.0                    & 26.0                    & 10.0                   & 6.0                      & 0      & - & -                \\
                           & CoT   &4        & 76.0                    & 70.0                    & 58.0                    & 42.0                    & 30.0                    & 26.0                     & 20.0               & - & -          \\
                           & LtM &4  & 76.0                    & 72.0                    & 66.0                   & 50.0                   & 46.0                    & 36.0                     & 25.0               & - & -          \\
                           & SKiC  &2 & \textbf{81.0}           & \textbf{97.0}           & \textbf{77.0}           & \textbf{59.0}           & \textbf{56.0}           & \textbf{48.0}            & \textbf{36.0}            & - & -    \\  \midrule
\multirow{5}{*}{text-davinci-003}  & zero-shot   &0        & 0                     & 0                     & 0                     & 0                     & 0                     & 0                      & 0                     & - & -     \\
                           & 4-shots  &4     & 99.0                    & 97.0                    & 89.0                    & 68.0                    & 45.0                    & 27.0                     & 10.0            & - & -             \\
                           & CoT  &4        & 100.0                   & 99.0                    & 90.0                    & 75.0                    & 52.0                    & 39.0                     & 31.0                    & - & -     \\
                           & LtM &4 & 100.0                   & 99.0                    & 94.0                    & 90.0                    & 87.0                    & 84.0                     & 80.0                   & - & -      \\
                           & SKiC  &2 & \textbf{100.0}          & \textbf{100.0}          & \textbf{100.0}          & \textbf{100.0}          & \textbf{100.0}          & \textbf{99.0}           & \textbf{98.0}          & - & -      \\  \midrule
\multirow{5}{*}{ChatGPT}   & zero-shot &0           & 99.0                    & 98.0                    & 93.0                    & 88.0                    & 84.0                    & 80.0                     & 77.0                      & 38.0 & 16.0   \\
                           & 4-shots  &4      & 100.0                   & 100.0                   & 95.0                    & 92.0                    & 90.0                    & 86.0                     & 85.0                    & 46.0 & 28.0     \\
                           & CoT    &4       & 100.0                   & 100.0                   & 97.0                    & 95.0                    & 92.0                    & 88.0                     & 85.0                    & 62.0 & 56.0     \\
                           & LtM &4 & 100.0                   & 100.0                   & 99.0                    & 95.0                    & 92.0                    & 92.0                     & 88.0            & 80.0 & 76.0          \\
                           & SKiC  &2 & \textbf{100.0}          & \textbf{100.0}          & \textbf{100.0}          & \textbf{100.0}          & \textbf{100.0}          & \textbf{100.0}           & \textbf{100.0}        & \textbf{100.0} & \textbf{100.0}  \\ \bottomrule    
\end{tabular}
}
\end{table*}

\begin{table*}[t]
\caption{Accuracy on the task of adding and multiplying two numbers with different digits. For the addition or multiplication task, the exemplars include how to add or multiply two numbers with 2 or 3 digits. Therefore, the results for adding numbers with $4 \sim 7$ digits and multiplying numbers with 4 and 5 digits measure the compositional generalization capabilities over harder problems. We also compare GPT3 finetuned with scratchpad method \citep{dziri2023faith} on the multiplication task.} \label{Tab:add_mul_results}
\centering
\scalebox{0.75}{
\begin{tabular}{c|c|c|cc|cccc||cc|cc} \toprule
\multirow{2}{*}{\textbf{Model}}     & \multirow{2}{*}{\textbf{Prompting}} &\multirow{2}{*}{\textbf{\#-shots}} & \multicolumn{6}{|c||}{\textbf{Addition}} & \multicolumn{4}{|c}{\textbf{Multiplication}}   \\  \cmidrule{4-13}

 & & &\multicolumn{1}{c}{\textbf{2}} & \multicolumn{1}{c|}{\textbf{3}} & \textbf{4}   & \textbf{5}   & \textbf{6}   & \textbf{7} &\multicolumn{1}{c}{\textbf{2}} & \multicolumn{1}{c|}{\textbf{3}} & \textbf{4}   & \textbf{5} \\ \midrule \midrule
\multirow{4}{*}{LLAMA-65B} & zero-shot    &0              & 58.0                             & 40.5                             & 22.5           & 8.0            & 0            & 0            & 28.0                             & 17.0                             & 0                              & 0                              \\
                           & 4-shots  &4            & 64.5                             & 46.5                             & 28.0           & 10.0            & 0            & 0            & 24.0                             & 18.0                             & 0                              & 0                              \\
                           & CoT   &4             & 60.0                             & 52.5                             & 24.0           & 12.0           & 1.0            & 0           & 22.0                             & 21.0                             & 0                              & 0                              \\
                           & SKiC &2     & \textbf{82.5}                    & \textbf{74.5}                    & \textbf{66.5}  & \textbf{52.0}  & \textbf{38.0}  & \textbf{22.0}   & \textbf{50.0}                    & \textbf{42.0}                    & \textbf{12.0}                    & \textbf{8.0}                     \\   \midrule
\multirow{5}{*}{text-davinci-003}  & zero-shot  &0                & 100.0                            & 100.0                            & 98.0           & 87.5           & 74.5           & 54.0           & 76.0                             & 14.5                             & 0                              & 0                              \\
                           & 4-shots &4               & 100.0                            & 100.0                            & 98.0           & 92.0           & 80.5           & 58.5           & 82.0                             & 18.0                             & 0                              & 0                              \\
                           & CoT    &4               & 100.0                            & 100.0                            & 92.0           & 68.5           & 42.0           & 38.0            & 86.0                             & 20.5                             & 2.0                              & 0                              \\ 
                             & finetuned &0 &- &- &- & - &- & -& 99.0 &55.0 & 1.0  &0.0 \\ 
                           & SKiC  &2   & \textbf{100.0}                   & \textbf{100.0}                   & \textbf{99.0}  & \textbf{98.0}  & \textbf{99.0}  & \textbf{98.5}   & \textbf{100.0}                   & \textbf{58.0}                    & \textbf{42.5}                    & \textbf{36.0}                    \\  \midrule
\multirow{5}{*}{ChatGPT}   & zero-shot  &0                & 100.0                            & 100.0                            & 100.0          & 92.0           & 86.5           & 78.0           & 99.0                            & 55.0                            & 1.0                             & 0                              \\
                           & 4-shots   &4             & 100.0                            & 100.0                            & 100.0          & 94.0           & 90.5           & 83.5           & 99.0                             & 58.0                             & 1.0                              & 0                              \\
                           & CoT     &4              & 100.0                            & 100.0                            & 98.5           & 90.0           & 87.5           & 80.0           & 99.0                             & 54.5                             & 13.0                              & 2.0                              \\
                           & Algorithm  &2 &100,0 &100,0 & 98.0 & 94.5 & 91.5 & 90.0 & 100.0 & 68.0 & 20.0 & 0 \\
                           & SKiC &2      & \textbf{100.0}                   & \textbf{100.0}                   & \textbf{100.0} & \textbf{100.0} & \textbf{100.0} & \textbf{100.0} & \textbf{100.0}                   & \textbf{82.0}                    & \textbf{72.0}                    & \textbf{48.5}                    \\ \bottomrule
\end{tabular} 
}
\end{table*}

\begin{table*}[t]
\begin{minipage}[t]{0.36\textwidth}
\caption{Exact Match on Commaqa-E. The ``Comp. Gen'' column reports the results on the unseen questions from the compositional split.} \label{Tab:qa_results}
\centering
\scalebox{0.55}{
\begin{tabular}{c|c|c|cc} \toprule
\textbf{Model}             & \textbf{Prompting} &\textbf{\#-shots} & \multicolumn{1}{l}{\textbf{Test}} & \multicolumn{1}{l}{\textbf{Comp. Gen}} \\ \midrule \midrule
\multirow{5}{*}{LLAMA-65B} & zero-shot  &0             & 12.0                              & 16.3                                   \\
                           & 4-shots   &4          & 15.0                              & 24.6                                 \\
                           & CoT   &4             & 27.0                              & 30.8                                  \\
                           & Decomp  & 12           & 32.0                              & 40.4                                    \\
                           &  SKiC$\dag$ &2    & \textbf{44.0}                     & \textbf{52.0}                         \\ \midrule
\multirow{5}{*}{text-davinci-003}  & zero-shot  &0         & 34.0                              & 26.8                                   \\
                           & 4-shots    &4         & 42.0                              & 33.5                                  \\
                           & CoT      &4          & 44.0                              & 38.2                                 \\
                           & Decomp   &12          & 58.0                              & 66.6                                 \\
                           &  SKiC$\dag$   &2  & \textbf{66.0}                     & \textbf{74.8}                         \\ \midrule
\multirow{5}{*}{ChatGPT}   & zero-shot   &0         & 42.0                              & 30.6                                  \\
                           & 4-shots     &4        & 47.0                              & 40.3                                  \\
                           & CoT        &4        & 55.0                              & 46.4                                  \\
                           & Decomp    &12         & 64.0                              & 73.5                                  \\
                           & SKiC$\dag$   &2   & \textbf{70.0}                     & \textbf{80.8}                 \\ \bottomrule        
\end{tabular} 
}
\end{minipage}
\hfill
\begin{minipage}[t]{0.62\textwidth}
\caption{Accuracy on the dynamic programming task. The in-context exemplars are with sequence lengths of 4, 5. So the results for 6,7,8 measures the out-of-distribution generalization to harder problems. We also compare the finetuned text-davinci-003 with scratchpad.} \label{Tab:dp_results}
\centering
\scalebox{0.615}{
\begin{tabular}{c|c|c|cc|ccc} \toprule
\textbf{Model}               & \textbf{Prompting}&\textbf{\#-shots} & \multicolumn{1}{c}{\textbf{4}} & \textbf{5}  & \textbf{6}  & \textbf{7}  & \textbf{8}  \\ \midrule\midrule 

\multirow{5}{*}{text-davinci-003} & zero-shot  &0        & 10.5                              & 4.0           & 4.0           & 0.0           & 0.0           \\
                          & 4-shots     &4       & 32.5                             & 18.0          & 10.0          & 4.0           & 0.0           \\
                          & CoT     &4           & 58.0                             & 22.0          & 15.0          & 8.0           & 2.0           \\
                          &finetuned & 0 &\textbf{100.0} &\textbf{100.0} &22.0 &14.0 &8.0\\
                          & SKiC   &2             & 78.0                    & 62.5 & \textbf{54.5} & \textbf{48.0} & \textbf{42.5} \\ \midrule
\multirow{4}{*}{ChatGPT}  & zero-shot   &0       & 18.0                             & 10.0          & 6.0           & 4.0           & 0.0           \\
                          & 4-shot    &4         & 44.5                            & 18.0          & 10.0          & 4.0           & 0.0           \\
                          & CoT    &4            & 82.5                             & 76.0          & 72.0        & 64.0           & 55.5           \\
                          & SKiC   &2            & \textbf{98.0}                    & \textbf{96.0} & \textbf{95.0} & \textbf{94.0} & \textbf{92.0} \\ \midrule
\multirow{4}{*}{GPT4}     & zero-shot   &0       & 58.0                             & 42.5          & 35.5          & 28.0          & 12.0          \\
                          & 4-shots     &4       & 76.5                             & 70.5          & 58.0          & 55.0          & 42.0          \\
                          & CoT       &4         & 94.0                             & 91.0         & 88.0          & 83.5          & 72.0          \\
                          & SKiC  &2             & \textbf{100.0}                    & \textbf{100.0} & \textbf{100.0} & \textbf{99.0} & \textbf{98.0} \\ \bottomrule
\end{tabular} 
}
\end{minipage}
\end{table*}

\section{Comparison to Tool-Using Works} \label{sec:tool}

The major contribution of our work is to understand and unlock the \emph{inherent} composition abilities (easy-to-hard generalization) in LLMs themselves. The line of tool-using work is complementary with our work and can be easily integrated to substitute several basic skills to further improve the performances; that is, the external tools can also be interpreted as basic skills that the model can tap into. However, we focus only on how to tap into the internal basic skills for compositional generalization. With the abundance of work on tool utilization with LLMs, there are still great merits in studying the composition of internal skills for several reasons.

First, external tools like programs might bring in extra latency during inferences as LLMs need to call multiple external functions when dealing with complex problems. As a result, if some of the foundational skills are available and reliable from internal knowledge of LLM, we should consider how to exploit them directly with one-stage through our SKiC. In addition, the external tools are generally pre-defined and implemented ahead of time with a clear boundary about what it can do and it cannot do. However, in the real open world setting, the abundant ambiguity of problem may make it hard to identify a clear boundary about which tool to call, leading to errors that may cascade to later stages. LLMs are strong and flexible in composing the internal knowledge and skills to solve complex problems. In such situations, it may have advantage to let LLMs flexibly use its own internal knowledge to solve such ambiguous problems.

Second, it is hard/impossible to enumerate all the needed external skills (external calls) in the context for complex tasks, which would lower down the generalization abilities if the models are taught to rely on provided external calls. So, our SKiC also encourages models to utilize their internal skills not provided in the context to solve complex tasks. 

What is more, tool-using cases are more focused on math-related reasonings or problems that can be converted into programming problems. However, not all the tasks can be improved by external tools (e.g., QA in our Table~\ref{Tab:qa_results}). Therefore, SKiC is more general to different types of tasks. Indeed, tool-use can actually be viewed as one basic skill that could be integrated into SKiC, so that LLMs can flexibly compose both internal skills and external tools in a hybrid manner for solving even more complex real problems, which we leave as a future work.

\begin{table*}[t]
 \caption{Accuracy of different models with our SKiC prompts on different evaluation subsets of the last-letter-concatenation task. The testing problems with 1 and 2 words are in-distribution evaluation, while the ones with $4\sim 12$ words are (harder) out-of-distribution evaluations. } \label{Tab:last_letter_results_llama2}
\centering
\small
\begin{tabular}{c|c|c|cc|ccccc} \toprule
\textbf{Model}                & \textbf{Prompting} &\textbf{\#-shots}     & \multicolumn{1}{c}{\textbf{1}} & \multicolumn{1}{c|}{\textbf{2}} & \multicolumn{1}{c}{\textbf{4}} & \multicolumn{1}{c}{\textbf{6}} & \multicolumn{1}{c}{\textbf{8}} & \multicolumn{1}{c}{\textbf{10}} & \multicolumn{1}{c}{\textbf{12}} \\ \midrule \midrule

text-davinci-003  
                           & SKiC  &2 & 100.0         & 100.0          & 100.0          & 100.0          & 100.0          & 99.0           & 98.0            \\  
ChatGPT   
                           & SKiC  &2 & \textbf{100.0}          & \textbf{100.0}          & \textbf{100.0}          & \textbf{100.0}          & \textbf{100.0}          & \textbf{100.0}           & \textbf{100.0}       \\ \midrule
LLAMA-65B
                           & SKiC  &2 & 81.0          & 97.0           & 77.0           & 59.0           & 56.0           & 48.0            & 36.0            \\ 
LLAMA2-70B & SKiC  &2 & 100.0         & 99.0          & 100.0          & 99.0          & 98.0          & 97.0           & 95.0            \\  
\bottomrule    
\end{tabular}
\end{table*}

\begin{table*}[t]
\caption{Accuracy of different models with our SKiC prompts on the task of adding two numbers with different digits (2,3,4,5,6,7). The prompting exemplars are constructed to demonstrate the addition between two numbers with 2 or 3 digits. Therefore, the results for adding numbers with $4 \sim 7$ digits measure the desirable compositional generalization capabilities over harder problems. $\dag$ denotes our method.} \label{Tab:add_results_llama2}
\centering
\begin{tabular}{c|c|c|cc|cccc} \toprule
\textbf{Model}             & \textbf{Prompting} &\textbf{\#-shots} & \multicolumn{1}{c}{\textbf{2}} & \multicolumn{1}{c|}{\textbf{3}} & \textbf{4}   & \textbf{5}   & \textbf{6}   & \textbf{7}   \\ \midrule \midrule

\multirow{1}{*}{text-davinci-003}  
                           & SKiC$\dag$  &2   & 100.0                   & 100.0                   & 99.0  & 98.0  & 99.0  & 98.5  \\  
\multirow{1}{*}{ChatGPT} 
                           & SKiC$\dag$ &2      & \textbf{100.0}                   & \textbf{100.0}                   & \textbf{100.0} & \textbf{100.0} & \textbf{100.0} & \textbf{100.0} \\ \midrule
\multirow{1}{*}{LLAMA-65B} 
                           & SKiC$\dag$  &2     & 82.5                    & 74.5                    & 66.5  & 52.0  &38.0  & 22.0  \\  
\multirow{1}{*}{LLAMA2-70B} 
                           & SKiC$\dag$  &2     & 83.0                    &78.0                    & 68.0  &55.0  & 40.0  & 25.0  \\ 

\bottomrule
\end{tabular} 
\end{table*}

\begin{table*}[t]
\caption{Accuracy of different models with our SKiC prompts on the task of multiplying two numbers with different digits (2,3,4,5). The prompting exemplars are constructed to demonstrate how to multiply two numbers with 2 or 3 digits. Therefore, the results for multiplying numbers with 4 and 5 digits  measure the compositional generalization capability over harder problems. $\dag$ stands for our method.} \label{Tab:mul_results_llama2}
\centering
\begin{tabular}{c|c|c|cc|cc} \toprule
\textbf{Models}            & \textbf{Prompting} &\textbf{\#-shots} & \multicolumn{1}{c}{\textbf{2}} & \multicolumn{1}{c|}{\textbf{3}} & \textbf{4}                     & \textbf{5}                     \\ \midrule  \midrule

\multirow{1}{*}{text-davinci-003}  
                           & SKiC$\dag$  &2             & 100.0                   &58.0                    & 42.5                    &36.0                    \\  
\multirow{1}{*}{ChatGPT}  
                           & SKiC$\dag$  &2              & \textbf{100.0}                   & \textbf{82.0}                    & \textbf{72.0}                    & \textbf{48.5}                    \\ \midrule

\multirow{1}{*}{LLAMA-65B} 
                           & SKiC$\dag$    &2            & 50.0                    & 42.0                    & 12.0                    & 8.0                     \\ 

\multirow{1}{*}{LLAMA2-70B} 
                           & SKiC$\dag$    &2            & 99.0                  & 51.0                    & 15.0                    & 6.0                    \\ 

\bottomrule
                     
\end{tabular} 
\end{table*}

\begin{table*}[t]
\caption{Performance of different models with our SKiC prompts on Commaqa-E datasets (measured in Exact Match). The column of ``Comp. Gen'' reports the results on the new (unseen) compositional questions from the compositional generalization split. $\dag$ denotes our method.} \label{Tab:qa_results_llama2}
\centering
\begin{tabular}{c|c|c|cc} \toprule
\textbf{Model}             & \textbf{Prompting} &\textbf{\#-shots} & \multicolumn{1}{l}{\textbf{Test}} & \multicolumn{1}{l}{\textbf{Comp. Gen}} \\ \midrule \midrule

\multirow{1}{*}{text-davinci-003} 
                           &  SKiC$\dag$   &2  &66.0                     & 74.8                         \\ 
\multirow{1}{*}{ChatGPT}  
                           & SKiC$\dag$   &2   & \textbf{70.0}                    & \textbf{80.8}                 \\ \midrule

\multirow{1}{*}{LLAMA-65B} 
                           &  SKiC$\dag$ &2    & 44.0                     & 52.0                         \\ 

\multirow{1}{*}{LLAMA2-70B} 
                           &  SKiC$\dag$ &2    &46.7                  & 55.9                         \\ 
\bottomrule        
\end{tabular} 
\end{table*}

\begin{table*}[t]
\caption{Accuracy of different models with our SKiC prompts on the dynamic programming task with input sequence lengths being 4,5,6,7,8, respectively. The in-context exemplars for all the prompts are constructed with sequence lengths of 4 and 5. Therefore, the results for sequence lengths of 6,7,8 measures the out-of-distribution generalization to increasingly harder problems. $\dag$ denotes our method.} \label{Tab:dp_results_llama2}
\centering
\begin{tabular}{c|c|c|cc|ccc} \toprule
\textbf{DP}               & \textbf{Prompting}&\textbf{\#-shots} & \multicolumn{1}{c}{\textbf{4}} & \textbf{5}  & \textbf{6}  & \textbf{7}  & \textbf{8}  \\ \midrule\midrule 

\multirow{1}{*}{text-davinci-003}
                          & SKiC$\dag$   &2             & 78.0                    & 62.5 & 54.5 & 48.0 &42.5 \\ 
\multirow{1}{*}{ChatGPT} 
                          & SKiC$\dag$    &2            & 98.0                    & 96.0 & 95.0 & 94.0 & 92.0 \\ 
\multirow{1}{*}{GPT4}    
                          & SKiC$\dag$  &2             & \textbf{100.0}                    & \textbf{100.0} & \textbf{100.0} & \textbf{99.0} & \textbf{98.0} \\ \midrule

\multirow{1}{*}{LLAMA2-70B}
                          & SKiC$\dag$   &2             & 79.0                    & 78.0 & 70.0 & 68.0 &56.0 \\ 
\bottomrule
\end{tabular} 
\end{table*}

\section{Experimental Setup} \label{Sec:setup}
In this section, we explain our experimental settings in details. We show the superior compositional capabilities of our SKiC prompting by evaluating it in two settings:
\begin{itemize}[leftmargin=0.6cm]
  \item \textbf{Systematic Generalization: Composition over in-context skills}, where all the essential skills needed to solve the problems are provided in the context. The tasks we evaluate in this setting  include symbolic manipulation \citep{wei2022chain,zhou2022least,khot2022decomposed}, arithmetic operation \citep{dziri2023faith}, question answering \citep{khot2022decomposed}, and dynamic programming \citep{dziri2023faith}. In this setting, we mainly examine the ability to generalize from easy demonstration exemplars to more difficult testing problems (i.e., easy-to-hard generalization).
  
  \item \textbf{Enhanced Complex Reasoning: Generalization beyond in-context skills}, where models also need to harness skills beyond what have been provided in the context and tap into the internal skills for math reasoning like GSM8K~\citep{wei2022chain,zhou2022least} and MATH~\citep{hendrycks2021measuring} problems. In this context, the challenge lies in achieving diverse compositions across a wide range of foundational skills for complex reasoning.
\end{itemize}

\subsection{Systematic Generalization: Composition over In-Context Skills}
We begin by evaluating SKiC on tasks that require only a limited skill set, yet pose challenges in terms of easy-to-hard generalization capabilities. Under these circumstances, we construct our SKiC prompts manually, adhering to the first methodology outlined in Appendix~\ref{sec:more_details_construction}. We mainly consider foundation models including LLAMA-65B \citep{touvron2023llama1}, text-davinvi-003 \citep{brown2020language}, ChatGPT and GPT4~\citep{openai2023gpt4}. Additional experiments on LLAMA2~\citep{touvron2023llama} can be found in Appendix~\ref{Sec:llama2}.

\subsubsection{Symbolic Manipulation: Last Letters}
Following \citeauthor{zhou2022least}, we first assess the compositionality in LLMs through the last-letter-concatenation task. For a given list of words, the LLM needs to generate an output that is the concatenation of the last letter from each word in the list. We compare our SKiC with zero/few-shot standard prompting (4-shot) \citep{brown2020language}, CoT~\citep{wei2022chain} and Least-to-Most prompting (LtM) \citep{zhou2022least} on different large language models, including LLAMA-65B \citep{touvron2023llama1}, text-davinvi-003 \citep{brown2020language,ouyang2022training}, and ChatGPT. And we evaluate them on different subsets of testing problems that include 1, 2, 4, 6, 8, 10, 12, 50, 100 words\footnote{From \url{https://github.com/first20hours/google-10000-english/tree/master}.}, respectively. The exemplars in all the prompts are constructed from the cases with 1 or 2 words. Therefore, the evaluations on the test subsets with 1, 2 words are in-distribution, and the ones on 4, 6, 8, 10, 12 words are out-of-distribution. A SKiC prompt contains the skills and two examples of how to compose these skills as shown in Figure~\ref{Tab:last_letter_skill} and Figure~\ref{Tab:compose_last_letter_skill}. The model is given the needed skills such as putting the given words to a list and getting the last letter of one word, and then two examples of how to compose these skills to take the last letters of two given words.

\subsubsection{Arithmetic Operation}
Following \citeauthor{dziri2023faith}, we evaluate the compositional capabilities on two arithmetic operation tasks: addition and multiplication. These two tasks involves complicated composition over skills such as one-digit addition or multiplication, carry over, concatenation and etc.\citep{dziri2023faith}, making it difficult especially for long form addition or multiplication. We compare SKiC with zero/few-shot standard prompting \citep{brown2020language}, Chain-of-Thoughts prompting (CoT) \citep{wei2022chain}  and Algorithmic prompting \citep{zhou2022teaching} on different foundation models including LLAMA-65B, text-davinvi-003, and ChatGPT. We exclude the Least-to-Most prompting \citep{zhou2022least} as it is difficult to design linear problem decomposition for addition or multiplication task. We also include text-davinci-003 finetuned with scratchpad method \citep{nye2021show,dziri2023faith} on the multiplication task for comparison.

\paragraph{Addition} We construct different subsets of testing problems, which ask to output the sum of two numbers with 2,3,4,5,6,7 digits, respectively. The given in-context exemplars are only constructed to demonstrate the addition of two numbers with 2-digits or 3-digits. Consequently, the results for 4,5,6,7-digits summation are out-of-distribution evaluation. We show SKiC prompting for the addition task in Figures~\ref{Tab:simple_add_skill}-\ref{Tab:compose_simple_add_skill}, which show the skills and one compositional exemplar, respectively. We first present the basic skills (e.g., extracting digits from a number) and then show how to use these skills to add two numbers with two examples.

\paragraph{Multiplication} 
Next, we evaluate the compositional generalization performance on the multiplication task. Specifically, we construct different subsets of evaluation problems that ask for the product of two numbers with 2,3,4,5 digits, respectively. The given in-context exemplars in all the prompts are constructed to demonstrate 2-digit and 3-digit multiplications. Therefore, the results for 4,5-digits multiplications measure the compositional generalization to unseen harder problems.
The construction of our Skills-in-Context prompting is shown in Figure~\ref{Tab:simple_mul_skill} and Figure~\ref{Tab:compose_simple_mul_skill}, which illustrate the skills and the compositional exemplar, respectively.

\subsubsection{Long-Context Question Answering: CommaQA-E}  
To evaluate the compositional generalization in the reading comprehension setting, following \citeauthor{khot2022decomposed}, we evaluate different prompting strategies on CommaQA-E~\citep{khot2021hey}. For given facts of a set of synthetically generated entities, the models need to answer the multi-hop questions which are composed of multiple reasoning steps, e.g., \textit{What movies have people from the country Stridery acted in?}. Besides the standard zero/few-shot prompting \citep{brown2020language} and the Chain-of-Thoughts prompting (CoT) \citep{wei2022chain}, we also compare our SKiC prompting to Decomp prompting\footnote{Reproduced using the original code from: \url{https://github.com/allenai/DecomP/tree/main}} \citep{khot2022decomposed}. We evaluate the results on different foundation models: LLAMA-65B, text-davinvi-003, and ChatGPT. The construction of the SKiC prompting for CommaQA-E is described in Figures~\ref{Tab:qa_skill}-\ref{Tab:compose_qa_skill}, which show the skills and the exemplars of how to compose the skills, respectively. Notably, both the ability to break down complex questions into simple ones and the operation to answer each simple questions are also treated as (basic) skills --- see Figure \ref{Tab:qa_skill}.

\subsubsection{Dynamic Programming}
We then further evaluate the compositional generalization capabilities of SKiC in solving a classic dynamic programming problem \citep{dziri2023faith}: \textit{Given a sequence of integers, find a subsequence with the highest sum, such that no two numbers in the subsequence are adjacent in the original sequence.} We compare our SKiC prompting with standard zero/few-shot prompting \citep{brown2020language}, and Chain-of-Thoughts prompting (CoT)\footnote{The reasoning steps are constructed based on the scratchpad prompts used in \citet{dziri2023faith}.}  \citep{wei2022chain} on different LLMs (text-davinvi-003, ChatGPT and GPT4). In addition, we also compare with the baseline of finetuned text-davinci-003 with scratchpad from \citep{dziri2023faith}.
Likewise, we evaluate them on different subsets of testing instances with sequence length of 4, 5, 6, 7, 8, respectively.\footnote{The numbers are within the range [-5,5]} The in-context exemplars are constructed with sequence length of 4 and 5. Therefore, the testing subsets with sequence length of 4 and 5 are in-distribution evaluation and the ones with length 6, 7, and 8 are for out-of-distribution evaluation. The construction of SKiC is characterized in Figures~\ref{Tab:dp_skill}-\ref{Tab:compose_dp_skill}, which show the skills and their compositions exemplars, respectively. Specifically, in the SKiC prompt, the models are presented with the skills to get the length of a list, find the max number for a given list and add two single digit numbers, followed by two compositional exemplars about how to compose these skills to solve the dynamic programming problems with sequence lengths being 4 and 5.

\subsection{Enhanced Complex Reasoning: Generalization Beyond In-Context Skills}
We further evaluate whether our SKiC prompting could allow LLMs to generalize beyond the skills provided in the prompt context and invoke the massive set of internal skills and knowledge that are acquired during pre-training. 
Such capability is vital in solving complex reasoning problems (e.g., math), which require varied compositions over a vast amount of foundational skills. And it is impractical to enumerate all the skills in context.

\subsubsection{GSM8K}
We first apply our SKiC prompting to GSM8K \citep{cobbe2021training}, which requires multiple math-related skills to solve complex math world problems. We construct our SKiC prompt by using the first approach in Appendix~\ref{sec:more_details_construction}, which includes a limited skill set together with eight compositional exemplars to teach the LLMs how to use them. Figures~\ref{Tab:gsm8k_skill}-\ref{Tab:compose_gsm8k_skill} show the constructed skill set and one compositional exemplar, respectively. We compare our SKiC with Chain-of-Thoughts prompting (CoT) \citep{wei2022chain}, Least-to-Most prompting (LtM) \citep{zhou2022least}, ComplexCot \citep{fu2022complexity} and PHP \citep{zheng2023progressive} on different foundation models (i.e., text-davinvi-003, ChatGPT and GPT-4).

\subsubsection{MATH}
We then apply SKiC to MATH~\citep{hendrycks2021measuring}, which is a significantly more challenging benchmark on mathematical reasoning. It encompasses problems in Algebra, Counting and Probability, Geometry, Intermediate Algebra, Number Theory, PreAlgebra, and PreCalculus. Due to the large variety of foundational capabilities needed for solving these math problems, it is infeasible to distill and enumerate the needed skills manually. Therefore, we adopt the second approach as described in Appendix~\ref{sec:more_details_construction}, where we prompt the LLM to generate the skills and then craft the compositional examples manually. Specifically, we first prompt the LLM (i.e., the same LLM that we will use to solve the problems) to generate a list of skills for each subject category in the MATH dataset (e.g., ``Counting and Probability'') with the instruction ``Basic skills in $[$subject$]$''. Then we further ask the model to generate the description of each skill, and the resulting skill set is listed in Figure~\ref{Tab:math_skill}. In Figure~\ref{Tab:compose_math_skill}, we show a compositional exemplar that demonstrates how to utilize the skills to solve a problem in MATH dataset. Note from this example that we ground a part of the reasoning steps to in-context skills such as ``Combination'' and ``Sub'' and anchor others to internal skills (e.g., ``Pascal's Triangle''). In our experiment, we provide the model with seven exemplars (one example per category in the MATH dataset). We compare our SKiC prompting with different prompting strategies: CoT \citep{wei2022chain}, Scratchpad \citep{nye2021show}, Learning-to-Program(LtP) \citep{guo2023learning}, and ComplexCoT \citep{fu2022complexity} on two representative foundation models: ChatGPT and GPT-4 \footnote{We use the same model to construct the SKiC skills and to do the inference. That is, we prompt ChatGPT to construct the SKiC when testing with ChatGPT and we prompt GPT-4 to construct the SKiC when testing with GPT-4.}. In addition, we also include different ensemble strategies that are commonly combined together with these baselines: majority voting (maj1@k) \citep{lewkowycz2022solving}, Self-Consistency (SC) \citep{cot_wei_sc}, and Progressive-Hint Prompting (PHP) \citep{zheng2023progressive}.

\begin{table}[t]
 \caption{Accuracy of different sets of examples in CoT and our SKiC prompts on the last-letter-concatenation task with ChatGPT models.  } \label{Tab:different_prompts}
\centering
\small
\begin{tabular}{c|c|c} \toprule
\textbf{Examples in Prompts}                & \textbf{COT} &\textbf{SKiC}     \\  \midrule \midrule
'apple, banana'; 'apple, pie'    & 91.4 &\textbf{100.0} \\
 'math, code'; 'science, computer'   & 92.5 &\textbf{100.0} \\
 'ashc, edhoh'; 'shbod, wojois'     & 90.8 &\textbf{100.0} 
 \\   \bottomrule    
\end{tabular}
\end{table}

\begin{table}[t]
 \caption{Accuracy of different orders of examples in CoT and our SKiC prompts GSM8K task with ChatGPT models.  } \label{Tab:different_orders}
\centering
\begin{tabular}{c|c|c} \toprule
\textbf{Order of Examples}                & \textbf{COT} &\textbf{SKiC}     \\  \midrule \midrule
Order 1   & 74.4 &\textbf{87.2} \\
 Order 2   & 73.8 &\textbf{86.9} \\
Order 3    & 73.0 &\textbf{87.8} 
 \\   \bottomrule    
\end{tabular}
\end{table}

\begin{table}[t]
 \caption{Accuracy of MATH and FOLIO when using prompts designed for GSM8K with ChatGPT models.  } \label{Tab:different_tasks}
\centering
\small
\begin{tabular}{c|c|c} \toprule
\textbf{TASK}                & \textbf{COT for GSM8K} &\textbf{SKiC for GSM8K}     \\  \midrule \midrule
MATH    & 28.2 &\textbf{31.34} \\
FOLIO   & 68.8 &\textbf{72.5} \\   \bottomrule    
\end{tabular}
\end{table}

\section{Detailed Results for Systematic Generalization (Last Leter, Addition, Multiplication, Commaqa-E and DP)} \label{sec:detailed_results}

We report the results for last letter concatenation, addition\&multiplication, Commaqa-E and DP in Tables~\ref{Tab:last_letter_results}, ~\ref{Tab:add_mul_results}, ~\ref{Tab:qa_skill}, and ~\ref{Tab:dp_results}.

Standard zero/few-shot prompting generalizes poorly on the problems that are harder than the exemplars in the prompting context. For example, on last letter concatenation tasks, 4-shot standard prompting only achieves 10\% accuracy with text-davinci-003 when solving testing problems that involve 12 words. 
CoT, LtM and Decomp improve the overall performance but still degrade quickly over harder inputs (e.g., CoT slightly improves the accuracy on arithmetic tasks, LtM outperform CoT on last letter concatenation and Decomp prompting boosts the exact match on Commaqa-E dataset.). 
SKiC significantly boosts the performance with less demonstration examplesespecially in harder cases  (e.g., gaining over 68.9\% improvements on 7-digits summation with text-davinci-003 compared to baselines). Notably, SKiC achieves nearly perfect generalization on tasks like last letter concatenation, addition, and dynamic programming with text-davinci-003, ChatGPT or GPT4. 
Compared to fine-tuneded baselines such as finetuning text-davinci-003 with scratchpad, SKiC is also significantly better in the out-of-distribution regime, although its performance at the in-distribution regime is worse. \footnote{This is expected as the it is finetuned directly on input sequences with length 4 and 5, while our method is not finetuned at all.} 
These significant improvements demonstrate that by jointly presenting the models with skills and how to use these skills within the context, the models are instructed to resolve problems grounded to these basic skills. Consequently, it performs the reasoning steps more accurately and could generalize better to harder examples by following similar patterns to compose the basic skills. Examples of the generated answer with SKiC on these tasks when the inputs are harder can be found in Figures~\ref{Tab:example_last_letter_skill}--\ref{Tab:example_compose_dp_skill}.

\begin{figure*}[t]
\vskip 0.2in
\begin{center}
\centerline{\includegraphics[width=2.0\columnwidth]{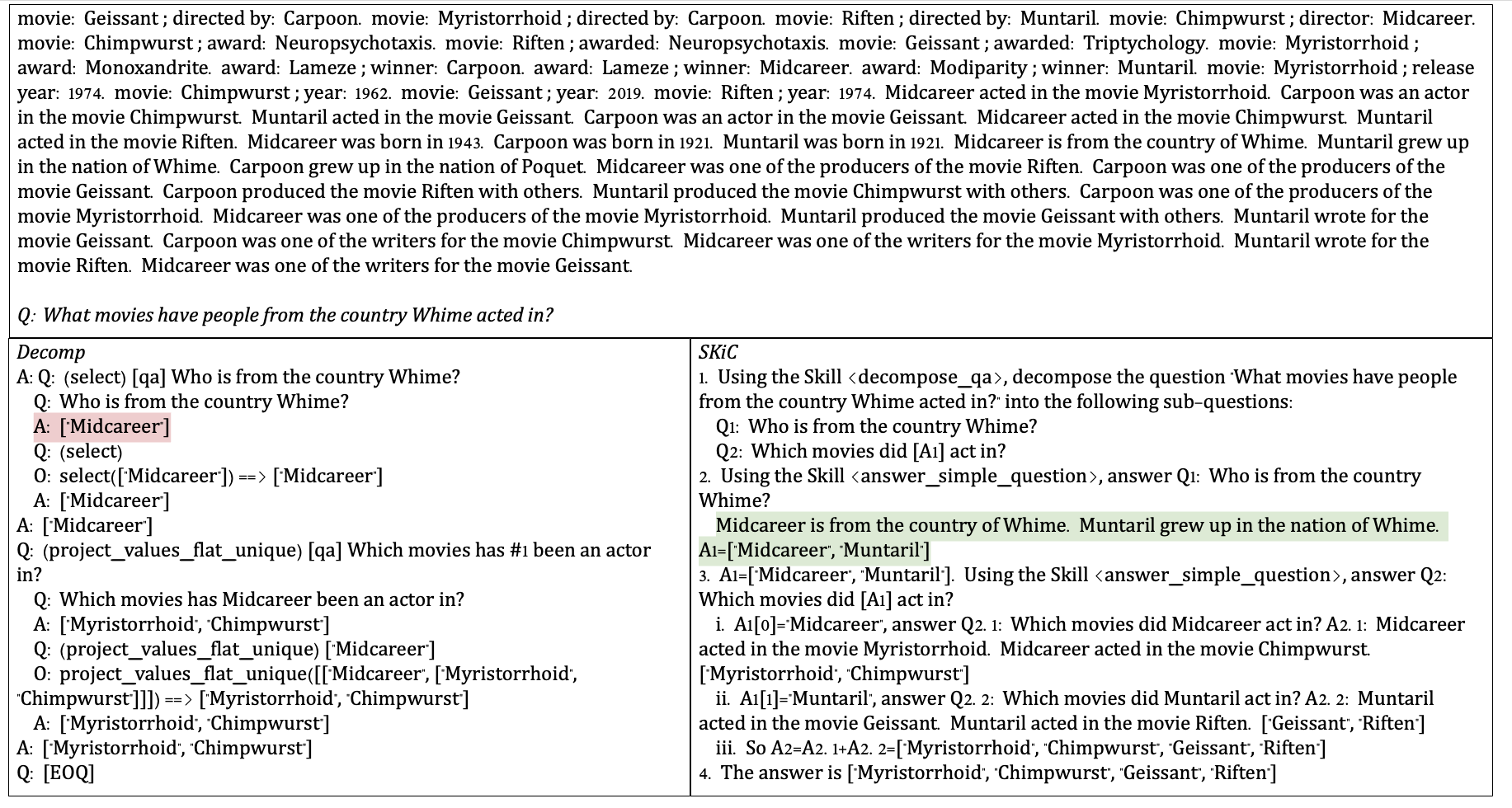}}
\caption{Examples of the generated answers from Decomp prompting and our SKiC prompting, respectively. The correct answer should be ``Myristorrhoid, Chimpwurst, Geissant, Riften''. Errors (highlighted in red) in early stages in Decomp propagate to final incorrect answers while our SKiC avoids such errors (highlighted in green).} 
\label{fig:error_example}
\end{center}
\vskip -0.2in
\end{figure*}

Results on Commaqa-E also illustrate the superiority of our 1-stage SKiC compared to multi-stage prompts. Unlike Decomp, both the ability to break down questions and answer simple questions are treated as skills in SKiC, and they are presented with the exemplars to demonstrate how to compose the skills (Figure~\ref{Tab:compose_qa_skill}) in the same context. Consequently, the LLM is able to flexibly apply these skills to reach the final answer within 1-stage, which could make different simple question answering help each other. For an example in Figure~\ref{fig:error_example}, errors made in early stages in Decomp result in wrong prediction while our SKiC accurately answer different questions in one context. This is a further manifestation of the advantage of concurrently demonstrating the skills and compositions.

\begin{table*}[t]
\caption{Accuracy and internal skill activation rate on the MATH with two different versions of SKiC on ChatGPT: the prompt with the skills generated from (i) ChatGPT and (ii) GPT-4. The \emph{internal skill activation rate} refers to the average proportion of skills utilized per question that originate from pre-trained knowledge (i.e., internal skills) rather than from the SKiC prompt context (i.e., the in-context skills).
} \label{Tab:math_ablation_results}
\centering
\scalebox{0.65}{
\begin{tabular}{c|c|ccccccc|c} \toprule
    \textbf{Metric}   &\textbf{Source of SKiC} & \textbf{Pre-Algebra}   & \textbf{Geometry}      & \textbf{Inter-Algebra} & \textbf{Algebra}       & \textbf{Probability}   & \textbf{Pre-Calculus}  & \textbf{NumTheory}     & \textbf{Overall}        \\  \midrule  \midrule

\multirow{2}{*}{Accuracy} &GPT4        & 60.7          & 27.8          & 16.8          & \textbf{58.2}          & 33.3          & 19.0          & 34.2          & 38.9          \\  

& ChatGPT  & \textbf{62.0} & \textbf{30.1} & \textbf{17.8} & 57.9 & \textbf{38.2} & \textbf{23.0} & \textbf{35.5} & \textbf{40.6} \\ \midrule 

\multirow{2}{*}{\begin{tabular}[c]{@{}c@{}}Internal Skill \\ Activation Rate\end{tabular}} &GPT4        & 5.9 & 18.5 & 11.2 & \textbf{6.6} & 7.0 &43.8 &6.2 & 12.5   \\   
        
    &ChatGPT               & \textbf{6.5} & \textbf{19.0} & \textbf{13.2} & 5.7 & \textbf{9.1} & \textbf{45.2} & \textbf{7.8} & \textbf{14.9} \\ \bottomrule
\end{tabular}
}
\end{table*}

\section{The Performance of SKiC on LLAMA2} \label{Sec:llama2}
We further evaluate the performance of SKiC prompting by using the LLAMA2 models \citep{touvron2023llama} on the following tasks: last latter concatenation, addition, multiplication, CommaQA-E, and dynamic programming tasks. The results are reported in the Tables~\ref{Tab:last_letter_results_llama2} and \ref{Tab:dp_results_llama2}. 

We observe that LLAMA2-70B generally outperforms LLAMA-65B and demonstrate stronger capabilities in following the exemplars for composing the in-context skills to solve the problems. There are still performance gaps between the open source LLAMA models and the proprietery LLMs such as text-davinci-003, ChatGPT and GPT4.


\begin{table}[t]
 \caption{Accuracy on Dynamic Programming task (8 numbers) of SKiC with ChatGPT after removing different components.} \label{Tab:different_components}
\centering
\begin{tabular}{c|c} \toprule
\textbf{Methods}                & \textbf{Dynamic Programming}   \\  \midrule \midrule
COT    & 72.0  \\ \midrule
SKiC    & \textbf{98.0}  \\ 
- in-context skill   & 94.0  \\
- skill grounding    & 82.0  \\ \bottomrule    
\end{tabular}
\end{table}

\begin{table}[t]
 \caption{Accuracy on SCAN with ChatGPT models.  } \label{Tab:scan}
\centering
\begin{tabular}{c|c} \toprule
\textbf{Methods}                & \textbf{SCAN}   \\  \midrule \midrule
COT &72.5 \\
SKiC & \textbf{100.0} 
 \\   \bottomrule    
\end{tabular}
\end{table}

\section{Different Sources of In-context Skills}
One important question we want to understand is whether it is beneficial to generate the in-context skills from the same foundation model used for prediction. Our hypothesis is that in-context skills generated from the same foundation model can initiate stronger synergize with the internal knowledge, due to their higher alignment. To test this hypothesis, we prompt the ChatGPT using the SKiC constructed from GPT-4 (i.e., the in-context skills are generated by GPT-4). The accuracy and the internal skill activation rate on MATH are reported in Table~\ref{Tab:math_ablation_results}. With the skills prompted from itself, we observe both improved accuracy and higher internal skill activation rate, even though the skills prompted from GPT-4 generally have higher quality. This suggests that (i) aligning the model that is used to prompt the in-context skills and the model for generating answers helps the models' capability to exploit internal skills, and (ii) activating more internal skills generally leads to higher performance, especially when solving problems that require compositions over wider range of skills.

\section{Robustness of Exemplars in SKiC}  \label{sec:ablation_prompts}

\paragraph{Different Choices of Exemplars}
We randomly selected exemplars in our SKiC prompts. The performance improvements are consistent even if we perturb the examples in the prompts. The results on last-letter tasks with ChatGPT with the use of different choices of few-shot exemplars in the prompts are shown in Table~\ref{Tab:different_prompts}. It shows the robustness of our proposed SKiC prompt to the selection of the few-shot exemplars.

\paragraph{Different Orders of Exemplars}
We also explore the order of different exemplars in the prompts. Through experiments, we find that the order of the examples also does not matter a lot because we randomly sample a limited number of examples (2 examples in most of the cases) to design SKiC. We shuffle the order in our prompts (consisting of 4 examples) for GSM8K and the performances are shown in Table~\ref{Tab:different_orders}.

\section{Generalization to New Tasks} \label{sec:generalization}
We further show that our SKiC which teach the model how to compose skills can also help the performances even if the provided prompts are designed for different tasks: We use the skills and prompts designed for GSM8K and directly apply them on MATH (competition-level math reasoning) \citep{hendrycks2021measuring} and FOLIO (logical inference) \citep{han2022folio} which are unseen tasks with ChatGPT as shown in Table~\ref{Tab:different_tasks}.

\begin{table*}[t]
 \caption{Accuracy on RTE and Last Letter with ChatGPT models.} \label{Tab:discover}
\centering
\begin{tabular}{c|c|c} \toprule
\textbf{Methods}                & \textbf{RTE} & \textbf{Last Letter (12 words)}  \\  \midrule \midrule
COT & 85.2 &72.5 \\ \midrule
SKiC &-  & \textbf{100.0}  \\
SKiC(Skills discoverd by LLM) & \textbf{89.8}  & \textbf{100.0} \\   \bottomrule    
\end{tabular}
\end{table*}

\begin{table*}[t]
 \caption{Accuracy on MATH for models fine-tuned with GSM8K data labeled with CoT reasoning steps and with SKiC reasoning steps. The one fine-tuned with SKiC reasoning steps show better weak-to-strong generalization.} \label{Tab:fine-tune}
\centering

\begin{tabular}{c|c|c|c} \toprule
\textbf{Model}             & \textbf{Train Set Source} &\textbf{Reasoning Step} & \multicolumn{1}{l}{\textbf{MATH}}  \\ \midrule \midrule
\multirow{3}{*}{LLAMA2-7B}  &- &- & 2.5 \\
                          & GSM8K & CoT         & 5.2    \\                         
                           & GSM8K & SKiC     & \textbf{7.6}                                            \\ \midrule
\multirow{3}{*}{LLAMA2-13B}  
                         &- &- & 3.9 \\
                          & GSM8K & CoT         & 5.1    \\                         
                           & GSM8K & SKiC     & \textbf{8.1}                                            \\\midrule
\multirow{3}{*}{LLAMA2-70B}  
                          &- &- & 13.5 \\
                          & GSM8K & CoT         & 14.1    \\                         
                           & GSM8K & SKiC     & \textbf{18.5}                                            \\ \bottomrule        
\end{tabular} 
\end{table*}

\section{Ablation of Different SKiC Components}  \label{sec:different_components}
Previous work \citep{khot2022decomposed,zhou2022least} introduced step-by-step reasoning and breaking down hard problems to simple problems to improve the easy-to-hard generalization. However, in our work,  we make another important discovery that, in order to teach models how to compose skills, it is also crucial to demonstrate the foundational skills and how to ground each of its reasoning steps onto the foundation skills. That is, besides step-by-step reasoning, explicit grounding is another key factor to elicit compositionality and easy-to-hard generalization. Our SKiC prompt structure constructed in this manner shows significantly better performances compared to previous work in all the experiments. Additionally, we perfrom ablation study to highlight our finding (the importance of skill grounding in reasoning steps). We compare SKiC with the setting where (i) we remove the skills but keep the skill grounding in reasoning steps and (ii) we remove the skill grounding in reasoning steps but keep the basic skill introduction in the front. The performance on Dynamic Programming is shown in Table~\ref{Tab:different_components}. Removing either part would bring in the performance drop, which further indicates the importance of skills and skill grounding in reasoning steps to improve the compositional generalization.

\section{Applying SKiC to Semantic Parsing} \label{Sec:scan}
We further design SKiC prompts and perform experiments on SCAN dataset \citep{chen2020compositional} that evaluates the ability to do semantic parsing. Specifically, our skills and examples of composing these skills are shown in Figures~\ref{Tab:scan_skill}-\ref{Tab:compose_scan_skill}. The performance with ChatGPT is shown in Table~\ref{Tab:scan}, which achieves perfect ($100\%$) performance.

\section{LLMs can automatically discover skills} \label{Sec:discover_skills}
We further provide experiments to show that the skills in our SKiC prompts can actually be discovered or summarized from examples by LLMs, which makes our SKiC more applicable to a wider range of tasks.  Specifically, we provide ChatGPT with 2 examples of NLI tasks from RTE~\citep{wang2018glue} and instruct ChatGPT to discover the skills from the given examples to perform the NLI tasks, which results in the skills including \textbf{Context Understanding} and \textbf{Inference Evaluation}. Based on the summarized skills from LLMs, we then construct our SKiC prompts and the results on RTE are shown in Table~\ref{Tab:discover}. Similarly, we utilize ChatGPT to discover skills for the last letter tasks which leading to the skill set including \textbf{Identify Words}, \textbf{Determine Last Letters}, \textbf{Concatenate Last Letters}, \textbf{Form New Sequence}. These are actually similar to what we have shown in Figure~\ref{Tab:last_letter_skill}. With such skills, we could further construct the SKiC prompts by adding these basic skills in the context and grounding reasoning steps onto these basic skills. This gives the similar performance compared to what we constructed manually as shown in Table~\ref{Tab:discover}. The results show the effectiveness of automatically discovering skills from LLMs and then using them to construct the SKiC prompts.

\section{SKiC Helps Instruction Tuning} \label{Sec:fine-tuning}
In this section, we show that instruction data which is constructed with SKiC can further be utilized to fine-tune LLMs to improve their capabilities of easy-to-hard generalization. Specifically, we generate training data by utilizing GPT4 to generate answers for GSM8K problems with SKiC prompts. That is, the generated reasoning steps for each GSM8K problem would be explicitly grounded to basic skills as shown in Figures~\ref{Tab:example_compose_gsm8k_skill_2}-\ref{Tab:example_compose_gsm8k_skill_3}. With the GSM8K data annotated with SKiC-format reasoning steps, we then finetune LLAMA2 models and evaluate their performances on MATH (which consists of significantly harder evaluation problems compared to the training problems from GSM8K) in zero-shot standard prompting settings. The results are shown in Table~\ref{Tab:fine-tune}. Compared to training data annotated with CoT reasoning steps, SKiC significantly improve the performances on MATH, which demonstrates that models that are fine-tuned with SKiC reasoning steps could achieve better generalization abilities to more complex and challenging testing cases. The results imply that SKiC data could potentially be used to replace CoT data in instruction tuning for eliciting stronger weak-to-strong generalization for LLMs.


\section{Generation Examples} \label{Sec:examples}
We further share some example generation from ChatGPT with our Skills-in-Context prompts on all the tasks in Figure~\ref{Tab:example_last_letter_skill},\ref{Tab:example_simple_add_skill}, \ref{Tab:example_simple_mul_skill}, \ref{Tab:example_compose_qa_skill}, \ref{Tab:example_compose_dp_skill}, \ref{Tab:example_compose_gsm8k_skill}, \ref{Tab:example_compose_gsm8k_skill_1}, \ref{Tab:example_compose_gsm8k_skill_2}, \ref{Tab:example_compose_gsm8k_skill_3}, \ref{Tab:example_compose_math_skill_1}, \ref{Tab:example_compose_math_skill_2}, \ref{Tab:example_compose_math_skill_3}, \ref{Tab:example_compose_math_skill_4}, \ref{Tab:example_compose_math_skill_5}.

\section{The Most Frequently Used Skills by GPT-4 for Solving MATH Benchmark}
In Table~\ref{Tab:math_top_skills}, we report the most frequently used skills by GPT-4 to solve the MATH problems. There are two sources of the skills: (i) the ones provided in the context of SKiC prompts, and (ii) the ones originating from GPT-4's internal knowledge (acquired through pretraining).

\begin{table*}[t]
\caption{The most frequently used skills by GPT-4 for solving MATH benchmark with SKiC prompting. The skills can be from the context of the SKiC prompts (denoted as ``in-context'' in the table) or from the internal knowledge acquired during the pretraining stage (denoted as ``internal'').} \label{Tab:math_top_skills}
\scalebox{0.72}{
\begin{tabular}{c|c|l} \toprule
\centering
\textbf{Category}              & \textbf{Source} & \textbf{Top Used Skills}                                                                                                                                                                                                            \\ \midrule \midrule
\multirow{3}{*}{Pre-Algebra}   & In-context            & Div, Mul, Add, Sub, Solve Equation, Area, Exp, Counting Principle, Radicals, Prime Numbers                                                                                                                                          \\ \cmidrule{2-3}
                               & Internal        & \begin{tabular}[c]{@{}l@{}}Pythagorean Theorem, Rounding, Divisibility Rules, Percentage, Angles, Simply Fraction, \\ Mean, Ratio, Triangle Angle Sum, Order of Operations\end{tabular}                                             \\  \midrule \midrule
\multirow{3}{*}{Geometry}      & In-context            & Area, Mul, Div, Add, Sub, Solve Equation, Volume, Radicals, Exp, Perimeter                                                                                                                                                          \\ \cmidrule{2-3}
                               & Internal        & \begin{tabular}[c]{@{}l@{}}Pythagorean Theorem, Trigonometry, Triangle, Triangle Inequality, Similar Triangles, \\ Circle, Geometry, Triangle Angle Sum, Angle Bisector Theorem, Trigonometric Ratios\end{tabular}                  \\ \midrule \midrule
\multirow{3}{*}{Inter-Algebra} & In-context            & Factoring, Solve Equation, Add, Mul, Sub, Complex Number, Inequality, Quadratic Formula, Div, Exp                                                                                                                                   \\ \cmidrule{2-3}
                               & Internal        & \begin{tabular}[c]{@{}l@{}}Substitution, Completing the Square, Polynomial, Logarithm, AM-GM Inequality, \\ Polynomial Division, Absolute Value, Summation, Sequences, Simplify\end{tabular}                                        \\ \midrule \midrule
\multirow{3}{*}{Algebra}       & In-context            & Add, Mul, Solve Equation, Sub, Div, Exp, Factoring, Quadratic Formula, Radicals, Distance Formula                                                                                                                                   \\ \cmidrule{2-3}
                               & Internal        & \begin{tabular}[c]{@{}l@{}}Absolute Value, Slope, Logarithm, Arithmetic Sequence, Completing the Square, Interval Notation, \\ Inverse Function, Substitution, Midpoint Formula, Ceiling Function\end{tabular}                      \\ \midrule \midrule
\multirow{3}{*}{Probability}   & In-context            & Factorial, Combination, Counting Principle, Probability, Add, Sub, Permutations, Mul, Div, Exp                                                                                                                                      \\ \cmidrule{2-3}
                               & Internal        & \begin{tabular}[c]{@{}l@{}}Simplify Fraction, Binomial Theorem, Expected Value, Arithmetic Sequence, Sum of Arithmetic Series, \\ Counting, Stars and Bars, Divisibility Rules, Binomial Probability, Perfect Squares\end{tabular}  \\ \midrule \midrule
\multirow{3}{*}{Pre-Calculus}  & In-context            & Solve Equation, Add, Mul, Sub, Complex Number, Div, Factoring, Radicals, Area, Distance Formula                                                                                                                                     \\ \cmidrule{2-3}
                               & Internal        & \begin{tabular}[c]{@{}l@{}}Trigonometric Identities, Trigonometry, Dot Product, Matrix Multiplication, Pythagorean Theorem, \\ Cross Product, Inverse Trigonometric Functions, Determinant, Vector Projection, Vectors\end{tabular} \\ \midrule \midrule
\multirow{3}{*}{NumTheory}     & In-context            & Add, Mod, Base Conversion, Mul, Congruences, Div, Sub, Factoring, Prime Number, GCD                                                                                                                                                 \\ \cmidrule{2-3}
                               & Internal        & \begin{tabular}[c]{@{}l@{}}Divisors, Divisibility Rules, Units Digit, Prime Fraction, Chinese Remainder Theorem, Arithmetic \\ Sequence, Exponents, Cyclic Patterns, Perfect Squares, Modular Arithmetic\end{tabular}    \\ \bottomrule           
\end{tabular}
}
\end{table*}

\newpage

\begin{strip}
\newpage
\begin{tcolorbox}[title = {Skills for Last Letter Concatenation}, colback = Apricot!25!white, colframe = BrickRed!75!black] 
Skill $<$words\_to\_list$>$: Put the asked words to a list.  For example, put the words in 'apple' to D=['apple']; put the words in 'apple, banana' to  D=['apple','banana']. 

\quad

Skill $<$last\_letter$>$: Get the last letter of one word.  For example, the last letter of 'apple' is 'e'; the last letter of 'banana' is 'a'.
\end{tcolorbox}
\noindent\begin{minipage}{\textwidth}
\captionof{figure}{The skills in Skills-in-Context prompt for last-letter-concatenation task.} \label{Tab:last_letter_skill}
\end{minipage}

\newpage

\begin{tcolorbox}[title = {An Example of Skill Composition for Last Letter Concatenation}, colback = Apricot!25!white, colframe = BrickRed!75!black] 
\textbf{Example}: Take the last letters of the words in 'apple, banana' and concatenate them.

\quad

\textbf{Answer}:

1. Using the Skill $<$words\_to\_list$>$, put the asked words, 'apple, banana', to a list. D=['apple','banana'] 

\quad

2. Get the last letter of each word in the list D=['apple','banana'] to a new list R=[]: 

\quad 
      \qquad i. Using the Skill $<$last\_letter$>$, the last letter of D[0]='apple' is 'e'. R=[e] 
      
     \qquad ii. Using the Skill $<$last\_letter$>$, the last letter of D[1]='banana' is 'a'. R=[e,a] 

\quad 

3. R=[e,a]. The answer is 'ea'.   
\end{tcolorbox}
\noindent\begin{minipage}{\textwidth}
\captionof{figure}{An exemplar of skill composition in Skills-in-Context prompt for last-letter-concatenation task.} \label{Tab:compose_last_letter_skill}
\end{minipage}

\newpage

\begin{tcolorbox}[title = {Skills for Addition}, colback = Apricot!25!white, colframe = BrickRed!75!black] 
Skill $<$extract\_digits$>$: Extract the digits in a number to a list.  \\
For example, Extract digits in 123 to D=[1,2,3]. Extract digits in 7654 to D=[7,6,5,4]. \\ \\

Skill $<$list\_length$>$: Get the number of elements in a list.  \\
For example, D=[1,2,3], len(D)=3. A=[1,2,4,5,6], len(A)=5. \\ \\

Skill $<$add\_two\_single\_digit\_number$>$: Add two single-digit numbers.  \\
0+0=0 0+1=1 0+2=2 0+3=3 0+4=4 0+5=5 0+6=6 0+7=7 0+8=8 0+9=9 \\
1+0=1 1+1=2 1+2=3 1+3=4 1+4=5 1+5=6 1+6=7 1+7=8 1+8=9 1+9=10 \\
2+0=2 2+1=3 2+2=4 2+3=5 2+4=6 2+5=7 2+6=8 2+7=9 2+8=10 2+9=11 \\
3+0=3 3+1=4 3+2=5 3+3=6 3+4=7 3+5=8 3+6=9 3+7=10 3+8=11 3+9=12 \\
4+0=4 4+1=5 4+2=6 4+3=7 4+4=8 4+5=9 4+6=10 4+7=11 4+8=12 4+9=13 \\
5+0=5 5+1=6 5+2=7 5+3=8 5+4=9 5+5=10 5+6=11 5+7=12 5+8=13 5+9=14 \\
6+0=6 6+1=7 6+2=8 6+3=9 6+4=10 6+5=11 6+6=12 6+7=13 6+8=14 6+9=15 \\
7+0=7 7+1=8 7+2=9 7+3=10 7+4=11 7+5=12 7+6=13 7+7=14 7+8=15 7+9=16 \\
8+0=8 8+1=9 8+2=10 8+3=11 8+4=12 8+5=13 8+6=14 8+7=15 8+8=16 8+9=17 \\
9+0=9 9+1=10 9+2=11 9+3=12 9+4=13 9+5=14 9+6=15 9+7=16 9+8=17 9+9=18 \\ \\

Skill $<$sub\_10$>$: Subtract 10 from a given number. \\
10-10=0 11-10=1 12-10=2 13-10=3 14-10=4 15-10=5 16-10=6 17-10=7 18-10=8 19-10=9 \\ \\

Skill $<$compare\_10$>$: Compare a number with 10.  \\
0$<$10 1$<$10 2$<$10 3$<$10 4$<$10 5$<$10 6$<$10 7$<$10 8$<$10 9$<$10 10$=$10 \\
11$>$10 12$>$10 13$>$10 14$>$10 15$>$10 16$>$10 17$>$10 18$>$10 19$>$10 
\end{tcolorbox}
\noindent\begin{minipage}{\textwidth} 
\captionof{figure}{The skills in Skills-in-Context prompt for the task of adding two numbers.} \label{Tab:simple_add_skill}
\end{minipage}

\newpage

\begin{tcolorbox}[title = {An Example of Skill Composition for Addition}, colback = Apricot!25!white, colframe = BrickRed!75!black] 
\textbf{Example}: Calculate 86+964: 

\quad

\textbf{Answer}:

1. Using Skill $<$extract\_digits$>$ and Skill $<$list\_length$>$, extract the digits in 86 to DM=[8,6]. len(DM)=2. Extract the digits in 964 to DN=[9,6,4]. len(DN)=3. \\ \\
2. len(DM)=2$<$len(DN)=3. Align the two lists by adding len(DN)-len(DM)=1 zeros to the shorter list (DM): DM=[0,8,6]. \\ \\
3. Using Skill $<$add\_two\_single\_digit\_number$>$, add every number in DM=[0,8,6] to the corresponding number in DN =[9,6,4] to form a list, DS=[0+9,8+6,6+4]=[9,14,10]. \\ \\
4. Append a 0 to the beginning of DS: DS=[0,9,14,10].Construct a list R=[0,0,0,0] with len(DS)=4 zeros.  \\ \\
5.  DS=[0,9,14,10] (DS[0]=0, DS[1]=9, DS[2]=14, DS[3]=10), starting from DS[3] to DS[1] (DS[3]=10, DS[2]=14, DS[1]=9): 

\quad 

    \qquad i. DS[3]=10, R[3]=0, R[3]=R[3]+DS[3]=0+10=10. Based on Skill $<$compare\_10$>$, R[3]=10=10, so R[2]=1 and R[3]=10-10=0 by Skill $<$sub\_10$>$. R=[R[0],R[1],R[2],R[3]]=[0,0,1,0].

    \qquad ii. DS[2]=14, R[2]=1, R[2]=R[2]+DS[2]=1+14=15. Based on Skill $<$compare\_10$>$, R[2]=15$>$10, so R[1]=1 and R[2]=15-10=5 by Skill $<$sub\_10$>$. R=[R[0],R[1],R[2],R[3]]=[0,1,5,0].

     \qquad iii. DS[1]=9, R[1]=1, R[1]=R[1]+DS[1]=1+9=10. Based on Skill $<$compare\_10$>$, R[1]=10=10, so R[0]=1 and R[1]=10-10=0 by Skill $<$sub\_10$>$. R=[R[0],R[1],R[2],R[3]]=[1,0,5,0]. 

\quad 

6. R=[1,0,5,0]. The answer is 1050.           
\end{tcolorbox}
\noindent\begin{minipage}{\textwidth}
\captionof{figure}{An exemplar of skill composition in Skills-in-Context prompting for the task of adding two numbers.} \label{Tab:compose_simple_add_skill}
\end{minipage}

\newpage

\begin{tcolorbox}[title = {Skills for Multiplication}, colback = Apricot!25!white, colframe = BrickRed!75!black] 
Skill $<$extract\_digits$>$: Extract the digits in a number to a list.  \\
For example, Extract digits in 123 to D=[1,2,3]. Extract digits in 7654 to D=[7,6,5,4]. \\ \\

Skill $<$list\_length$>$: Get the number of elements in a list.  \\
For example, D=[1,2,3], len(D)=3. A=[1,2,4,5,6], len(A)=5. \\ \\

Skill $<$mul\_two\_single\_digit\_number$>$: Multiply two single-digit numbers.  \\
0*1=0 0*2=0 0*3=0 0*4=0 0*5=0 0*6=0 0*7=0 0*8=0 0*9=0 \\
1*1=1 1*2=2 1*3=3 1*4=4 1*5=5 1*6=6 1*7=7 1*8=8 1*9=9 \\
2*1=2 2*2=4 2*3=6 2*4=8 2*5=10 2*6=12 2*7=14 2*8=16 2*9=18 \\
3*1=3 3*2=6 3*3=9 3*4=12 3*5=15 3*6=18 3*7=21 3*8=24 3*9=27 \\
4*1=4 4*2=8 4*3=12 4*4=16 4*5=20 4*6=24 4*7=28 4*8=32 4*9=36 \\
5*1=5 5*2=10 5*3=15 5*4=20 5*5=25 5*6=30 5*7=35 5*8=40 5*9=45 \\
6*1=6 6*2=12 6*3=18 6*4=24 6*5=30 6*6=36 6*7=42 6*8=48 6*9=54 \\
7*1=7 7*2=14 7*3=21 7*4=28 7*5=35 7*6=42 7*7=49 7*8=56 7*9=63 \\
8*1=8 8*2=16 8*3=24 8*4=32 8*5=40 8*6=48 8*7=56 8*8=64 8*9=72 \\
9*1=9 9*2=18 9*3=27 9*4=36 9*5=45 9*6=54 9*7=63 9*8=72 9*9=81 \\ \\

Skill $<$add\_multiple\_numbers$>$: Add multiple numbers such as m+n+p:

\quad 1. Add the first two numbers m+n and get the result r1=m+n. 

\quad 2. Add the third number p to r1 and get the result r2=r1+p.  

For example, calculate 128+64+79: 

\quad 1. Add the first two numbers 128+64 and get the result r1=128+64=192. 

\quad 2. Add the third number 79 to r1 and get the result r2=192+79=271. 

\quad So 128+64+79=271       
\end{tcolorbox}
\noindent\begin{minipage}{\textwidth}  
\captionof{figure}{The skills in Skills-in-Context prompt for the task of multiplying two numbers.} \label{Tab:simple_mul_skill}
\end{minipage}

\newpage

\begin{tcolorbox}[title = {An Example of Skill Composition for Multiplication}, colback = Apricot!25!white, colframe = BrickRed!75!black] 
\textbf{Example}: Calculate 184*67: 

\quad

\textbf{Answer}:

1. Using Skill $<$extract\_digits$>$ and Skill $<$list\_length$>$, extract the digits in 184 to DM=[1,8,4].  len(DM)=3. Extract the digits in 67 to DN=[6,7]. len(DN)=2. \\ \\
2. Add 0,1,len(DM)-1=2 zeros to the end of every number in DM=[1,8,4] according to the position of the number in DM: DMO=[1*100,8*10,4*1]=[100,80,4]. \\ \\
3. Add 0,len(DN)-1=1 zeros to the end of every number in DN=[6,7] according to the position of the number in DN: DNO=[6*10,7*1]=[60,7]. \\ \\
4. Using Skill $<$mul\_two\_single\_digit\_number$>$, multiple every number in DMO=[100,80,4] with every number in DNO=[60,7] and get R=[100*60,100*7,80*60,80*7,4*60,4*7]= [6000,700,4800,560,240,28]. \\ \\
5. Using Skill $<$add\_multiple\_numbers$>$, add all the numbers in R=[6000,700,4800,560,240,28], 6000+700+4800+560+240+28:

\quad 

    \qquad i. Add the first two numbers: r1=6000+700=6700. 
    
    \qquad ii. Add the third number 4800 to r1=6700: r2=6700+4800=11500. 
    
    \qquad iii. Add the fourth number 560 to r2=11500: r3=11500+560=12060. 
    
    \qquad iv. Add the fifth number 240 to r3=12060: r4=12060+240=12300. 
    
    \qquad v. Add the sixth number 28 to r4=12300: r5=12300+28=12328. 

\quad
    
6. So the answer is 12328 
\end{tcolorbox}
\noindent\begin{minipage}{\textwidth} 
\captionof{figure}{An exemplar of skill composition in Skills-in-Context prompting for the task of multiplying two numbers.} \label{Tab:compose_simple_mul_skill}
\end{minipage}

\newpage

\begin{tcolorbox}[title = {Skills for CommaQA-E task}, colback = Apricot!25!white, colframe = BrickRed!75!black] 
Skill $<$decompose\_qa$>$: Decompose a complex question into a set of sub-questions. \\ 
For example: Decompose the question "What awards have movies produced by people born \\ in 1910 won?"  
into the following sub-questions: \\
Q1: Who were born in the year 1910? \\
Q2: Which movies did [A1] produce? \\
Q3: Which awards were given to [A2]? \\
Decompose the question "What movies have people from the country Stridery acted in?" into  \\ 
the following sub-questions:
Q1: Who is from the country Stridery? \\
Q2: Which movies did [A1] act in? \\ \\

Skill $<$answer\_simple\_question$>$: Answer simple questions about the passage. \\
For example:  \\
$[$A Passage$]$ \\  \\
Q: Which awards were given to Zalate? \\
A: movie: Zalate ; awarded: Hallowcock. ["Hallowcock”] \\
Q: Which movies were given the Hallowcock award? \\
A: movie: Zalate ; awarded: Hallowcock. movie: SkirtSiCine ; award: Hallowcock. ["Zalate", \\ "SkirtSiCine"]  \\
Q: Which movies did Muntaril direct? \\
A: movie: Premercy ; directed by: Muntaril. ["Premercy] \\
Q: Which movies did Muntrail produce? \\
A: Muntaril produced the movie Premercy with others. Muntaril produced the movie SkirtSiCine  \\ 
with others. ["Premercy", "SkirtSiCine"] \\
Q: Which movies did Muntrail write? \\
A: Muntaril was one of the writers for the movie Zalate. Muntaril wrote for the movie \\
Featsaw. ["Zalate", "Featsaw"] \\
Q: Who are the actors in the movie Premercy? \\ 
A: Monsterscar was an actor in the movie Premercy. ["Monsterscar"] \\
Q: When was the moive Featsaw released? \\
A: Monsterscar was an actor in the movie Premercy. ["1973"] \\
\end{tcolorbox}
\noindent\begin{minipage}{\textwidth} 
\captionof{figure}{The skills in Skills-in-Context prompt for the CommaQA-E task.} \label{Tab:qa_skill}
\end{minipage}

\newpage

\begin{tcolorbox}[title = {An Example of Skill Composition for the CommaQA-E Task}, colback = Apricot!25!white, colframe = BrickRed!75!black] 
\textbf{Example}: $[$A Passage$]$ \\ 
Q: What movies have people from the country Stridery acted in? 

\quad

\textbf{Answer}:

1. Using Skill $<$decompose\_qa$>$, decompose the question "What movies have people from the country Stridery acted in?" into the following sub-questions:  

\quad

\qquad Q1: Who is from the country Stridery? 

\qquad Q2: Which movies did [A1] act in? 

\quad

2. Using Skill $<$answer\_simple\_question$>$, answer Q1: Who is from the country Stridery? 

\quad 

\qquad  Gastrat grew up in the nation of Stridery. A1=["Gastrat"] 

\quad 

3. A1=["Gastrat"], answer Q2: Which movies did [A1] act in? 

\quad

 \qquad   i. A1[0]="Gastrat", Using Skill $<$answer\_simple\_question$>$, answer Q2.1: Which movies did Gastrat act in?  

  \qquad \quad A2.1: Gastrat was an actor in the movie Partnershipmaker. Gastrat was an actor in the movie Partnershipmaker. Gastrat acted in the movie Warpstone.  A2.1=["Partnershipmaker", "Nilitude", "Warpstone"] 

\quad
  
  \qquad  ii.So A2=[A2.1]=["Partnershipmaker", "Nilitude", "Warpstone"] 

\quad
  
4. The answer is ["Partnershipmaker", "Nilitude", "Warpstone"]
\end{tcolorbox}
\noindent\begin{minipage}{\textwidth}    
\captionof{figure}{An exemplar of skill composition in Skills-in-Context prompting for the CommaQA-E task.} \label{Tab:compose_qa_skill}
\end{minipage}

\newpage

\begin{tcolorbox}[title = {Skills for Dynamic Programming}, colback = Apricot!25!white, colframe = BrickRed!75!black] 
Skill $<$list\_length$>$: Get the number of elements in a list.  \\
For example, D=[1,2,3], len(D)=3. A=[1,2,3,4,5,6], len(A)=6. \\ \\

Skill $<$find\_max$>$: Find the max number in a list. \\
For example, max(2,0,-1)=2. max(-3,-2,-1)=-1. max(5,-2,0)=5. max(0,-2,4)=4.  max(0,0,0)=0. max(-2,-1,0)=0. \\ \\

Skill $<$add$>$: Add two numbers. \\
For example, 1+2=3. -2+1=-1. 0+(-3)=-3. 1+(-1)=0. \\
\end{tcolorbox}
\noindent\begin{minipage}{\textwidth}
\captionof{figure}{The skills in Skills-in-Context prompt for the task of dynamic programming.} \label{Tab:dp_skill}
\end{minipage}

\newpage

\begin{tcolorbox}[title = {An Example of Skill Composition for the Dynamic Programming Task}, colback = Apricot!25!white, colframe = BrickRed!75!black] 
\textbf{Example}: Compose these skills to do the task: Given a sequence of integers, find a subsequence with the highest sum, such that no two numbers in the subsequence are adjacent in the original sequence.  

A=[3,2,1,5].

 \quad

\textbf{Answer}:

1. A=[3,2,1,5] (A[0]=3,A[1]=2,A[2]=1,A[3]=5). Using Skill $<$list\_length$>$, len(A)=4. Construct a list D=[0,0,0,0] with the len(A)=4 zeros. \\ \\
2. Update the last element in D. A[3]=5. Using Skill $<$find\_max$>$, max(5,0)=5. D[3]=5 and D=[0,0,0,5]. \\ \\
3. Update the second last element in D. A[3] = 5, A[2] = 1. Using Skill $<$find\_max$>$, max(5,1,0) = 5. D[2] = 5 and D=[0,0,5,5]. \\ \\ 
4. Starting from the third last (len(A)-2=2) element in D, update each element in D based  on A=[3,2,1,5] (A[0]=3,A[1]=2,A[2]=1,A[3]=5): 

\quad

 \qquad   i. D[3]=5, D[2]=5, A[1]=2. Using Skill $<$add$>$, A[1]+D[3]=2+5=7.  Using Skill $<$find\_max$>$, max(D[2],A[1]+D[3],0)=max(5,7,0)=7.  D[1]=7 and D=[0,7,5,5].

  \qquad  ii. D[2]=5, D[1]=7, A[0]=3. Using Skill $<$add$>$, A[0]+D[2]=3+5=8. Using Skill $<$find\_max$>$, max(D[1],A[0]+D[2],0)=max(7,8,0)=8. D[0]=8 and D=[8,7,5,5]. 

\quad
  
5.  D=[8,7,5,5]. The highest sum is D[0]=8. 
\end{tcolorbox}
\noindent\begin{minipage}{\textwidth}    
\captionof{figure}{An exemplar of skill composition in Skills-in-Context prompting for the dynamic programming task to find the highest sum of the subsequence.} \label{Tab:compose_dp_skill}
\end{minipage}

\newpage

\begin{tcolorbox}[title = {Skills for GSM8K}, colback = Apricot!25!white, colframe = BrickRed!75!black,fontupper =\fontsize{9pt}{8pt}\selectfont] 
Skill $<$extract\_digits$>$: Extract the digits in a number to a list. For example, extract digits in 123 to D=[1,2,3]. Extract digits in 7654 to D=[7,6,5,4] \\ \\

Skill $<$list\_length$>$: Get the number of elements in a list. For example, D=[1,2,3], len(D)=3. A=[1,2,4,5,6], len(A)=5. \\ \\

Skill $<$add\_two\_single\_digit\_number$>$: Add two single-digit numbers. For example, 0+0=0 0+1=1 0+2=2 0+3=3 0+4=4 0+5=5 0+6=6 0+7=7 0+8=8 0+9=9 \\ \\

Skill $<$sub\_two\_single\_digit\_number$>$: Subtract two single-digit numbers.  For example, 0-0=0 0-1=-1 0-2=-2 0-3=-3 0-4=-4 0-5=-5 0-6=-6 0-7=-7 0-8=-8 0-9=-9 \\ \\

Skill $<$sub\_10$>$: Subtract 10 from a given number. 10-10=0 11-10=1 12-10=2 13-10=3 14-10=4 15-10=5 16-10=6 17-10=7 18-10=8 19-10=9 \\ \\

Skill $<$add\_10$>$: Add 10 to a given number. -10+10=0 -9+10=1 -8+10=2 -7+10=3 -6+10=4 -5+10=5 -4+10=6 -3+10=7 -2+10=8 -1+10=9\\ \\

Skill $<$compare\_0$>$: Compare a number with 0.  10$>$0 9$>$0 8$>$0 7$>$0 6$>$0 5$>$0 4$>$0 3$>$0 2$>$0 1$>$0 0=0 -1$>$0 -2$>$0 -3$>$0 -4$>$0 -5$>$0 -6$>$0 -7$>$0 -8$>$0 -9$>$0 \\ \\

Skill $<$compare\_10$>$: Compare a number with 10.  0$<$10 1$<$10 2$<$10 3$<$10 4$<$10 5$<$10 6$<$10 7$<$10 8$<$10 9$<$10 10$=$10 11$>$10 12$>$10 13$>$10 14$>$10 15$>$10 16$>$10 17$>$10 18$>$10 19$>$10  \\ \\

Skill $<$mul\_two\_single\_digit\_number$>$: Multiply two single-digit numbers.  For example, 4*1=4 4*2=8 4*3=12 4*4=16 4*5=20 4*6=24 4*7=28 4*8=32 4*9=36 \\ \\

Skill $<$add\_multiple\_numbers$>$: Add multiple numbers such as m+n+p:  \\
\quad 1. Add the first two numbers m+n and get the result r1=m+n. \\ 
\quad 2. Add the third number p to r1 and get the result r2=r1+p.  \\ \\

Skill $<$add$>$: Use the skills to add two numbers. For example, calculate 86+964 $[$The steps to perform the add$]$ \\ \\

Skill $<$mul$>$: Use the skills to multiply two numbers. For example, calculate 86*964 $[$The steps to perform the multiplication$]$ \\ \\

Skill $<$sub$>$:  Use the skills to subtract a number from another number. For example, calculate 964-86 $[$The steps to perform the subtractraction$]$ \\ \\

Skill $<$age$>$: Describe the age of a person.  If a person is P years old, Q years ago, the person was P-Q years old.  If a person is P years old, in Q years, the person will be P+Q years old. If person A is P years old, person B is Q years old, and person A is R years older than person B, then P=Q+R. If person A is P years old, person B is Q years old, and person A is R years younger than person B, then P=Q-R. \\ \\

Skill $<$solve\_equation$>$: Solve an equation. When subtracting or adding the same number from both sides of the equation, the equation is still true. When moving a number from one side of the equation to the other side, the sign of the number changes. When moving a multiplication from one side of the equation to the other side, the sign of the multiplication changes. When moving a division from one side of the equation to the other side, the sign of the division changes. \\
For example, if the equation is 3x+5=7, move +5 to the other side, then 3x=7-5. Using the Skill $<$sub$>$, 3x=7-5=2. Move 3 to the other side, then x=2/3. \\ 
\end{tcolorbox}
\noindent\begin{minipage}{\textwidth} 
\captionof{figure}{The skills in Skills-in-Context prompt for GSM8K.} \label{Tab:gsm8k_skill}
\end{minipage}

\newpage

\begin{tcolorbox}[title = {An Example of Skill Composition for GSM8K}, colback = Apricot!25!white, colframe = BrickRed!75!black] 
\textbf{Example}:  Four years ago, Kody was only half as old as Mohamed. If Mohamed is currently twice 30 years old,  how old is Kody? 

 \quad

\textbf{Answer}:

1. Mohamed is currently twice 30 years old. Using the Skill $<$mul$>$, Mohamed is currently 30*2 = 60 years old. \\ \\
2. Using Skill $<$age$>$, four years ago, Mohamed was 4 years younger than now. Using the Skill $<$sub$>$,  Mohamed was 60-4 = 56 years old.  \\ \\
3. Four years ago, Kody was only half as old as Mohamed. Using the Skill $<$div$>$, Kody was 56/2 = 28 years old. \\ \\
4. Using Skill $<$age$>$, currently, Kody is 4 years older than four years ago. Using the Skill $<$add$>$, Kody is currently 28+4 = 32 years old. \\ \\
5. The answer is 32. 
\end{tcolorbox}
\noindent\begin{minipage}{\textwidth}  
\captionof{figure}{An exemplar of skill composition in Skills-in-Context prompting for GSM8K math problems.} \label{Tab:compose_gsm8k_skill}
\end{minipage}

\newpage

\begin{tcolorbox}[title = {Skills for MATH}, colback = Apricot!25!white, colframe = BrickRed!75!black,fontupper =\fontsize{7.5pt}{5pt}\selectfont] 
You have the knowledge of many skills, the following are some examples: \\
Skill $<$Add$>$: Add two numbers. For example, 128+987=1115.  \\ \\

Skill $<$Sub$>$: Subtract a number from another number. For example, 128-67=61.   \\ \\

Skill $<$Mul$>$: Multiply two numbers. For example, 128*76=9728.   \\ \\

Skill $<$Div$>$: Divide a number from another number. For example 12/3=4.  \\ \\

Skill $<$Mod$>$: Modulus or modulo, it finds the remainder of a division operation. For example, 10 mod 3 = 1, because 10 divided by 3 leaves a remainder of 1.  \\ \\

Skill $<$Exp$>$: An exponent refers to the number of times a number is multiplied by itself.  $[$More Details$]$ \\ \\

Skill $<$Base Conversion$>$: Base conversion is a way to change numbers from one base to another.   $[$More Details$]$\\ \\

Skill $<$Radicals$>$: A radical represents the root of a number. The square root (represented by sqrt) is the most common radical. $[$More Details$]$\\ \\

Skill $<$Factoring$>$: In the context of integers, factorization involves expressing a number as the product of prime numbers. $[$More Details$]$\\ \\

Skill $<$Solve Equation$>$: Solve an equation.   $[$More Details$]$\\ \\

Skill $<$Quadratic Formula$>$: The quadratic formula is used to solve quadratic equations. $[$More Details$]$\\ \\

Skill $<$Complex Number$>$: The quadratic formula is used to solve quadratic equations. $[$More Details$]$\\ \\

Skill $<$Piecewise Function: Continuous$>$: A piecewise function is continuous if it is continuous at every point in its domain. $[$More Details$]$\\ \\

Skill $<$Factorial$>$:  Factorial is a function that multiplies a given number by every number below it until 1. $[$More Details$]$\\ \\

Skill $<$Probability$>$:   Probability is the measure of the likelihood that an event will occur. $[$More Details$]$\\ \\

Skill $<$Conditional Probability$>$: The probability of an event occurring given that another event has already occurred.  $[$More Details$]$\\ \\

Skill $<$Probability Addition Rule$>$: The Addition Rule in probability is used to calculate the probability of either of two events happening. $[$More Details$]$\\ \\

Skill $<$Probability Multiplication Rule$>$: A way to determine the probability of two events occurring at the same time (conjointly). $[$More Details$]$\\ \\

Skill $<$Counting Principle$>$: If there are m ways to do one thing, and n ways to do another, then there are m*n ways of doing both. $[$More Details$]$\\ \\

Skill $<$Permutations$>$: Permutations refer to the arrangement of items in a specific order. $[$More Details$]$\\ \\

Skill $<$Combination$>$:  Combinations refer to the selection of items without regard to the order. $[$More Details$]$\\ \\

Skill $<$Perimeter$>$:  The perimeter of a shape is the distance around its boundary. $[$More Details$]$\\ \\

Skill $<$Area$>$:  The area of a shape is the amount of space that it covers. $[$More Details$]$\\ \\

Skill $<$Volume$>$:   Volume is the measure of the amount of space that a three-dimensional object occupies. $[$More Details$]$\\ \\

Skill $<$Prime Numbers$>$: A prime number is a natural number greater than 1 that has no positive divisors other than 1 and itself. $[$More Details$]$\\ \\

Skill $<$Composite Numbers$>$: Composite numbers are positive integers that have at least one positive divisor other than one or the number itself. $[$More Details$]$\\ \\

Skill $<$GCD$>$:The Greatest Common Divisor (GCD), also known as the Greatest Common Factor (GCF),  $[$More Details$]$\\ \\

Skill $<$LCM$>$:The Least Common Multiple (LCM) of two integers is the smallest positive integer that is divisible by both numbers without leaving a remainder. $[$More Details$]$\\ \\

Skill $<$Congruences$>$: Two integers a and b are said to be congruent modulo n if they have the same remainder when divided by n. $[$More Details$]$ \\
\end{tcolorbox}
\noindent\begin{minipage}{\textwidth} 
\captionof{figure}{The skills in Skills-in-Context prompt for MATH.} \label{Tab:math_skill}
\end{minipage}

\newpage

\begin{tcolorbox}[title = {An Example of Skill Composition for MATH}, colback = Apricot!25!white, colframe = BrickRed!75!black] 
\textbf{Example}: Shown below are rows 1, 2, and 3 of Pascal's triangle.$[$Pascal's triangle$]$. Let $(a_i),$ $(b_i),$ $(c_i)$ be the sequence, from left to right, of elements in the 2005th, 2006th, and 2007th rows, respectively, with the leftmost element occurring at $i = 0.$  Compute $[\sum_{i = 0}^{2006} \frac{b_i}{c_i} - \sum_{i = 0}^{2005} \frac{a_i}{b_i}.]$ \\ \\
 \quad

\textbf{Answer}:

1. Using the Skill $<$Pascal's Triangle$>$, the number in the n-th row and k-th column of the Pascal's triangle is C(n,k). \\ \\
2. Suppose $(a_i), (b_i), (c_i)$ represent the entries in rows n - 1, n, n + 1 of Pascal's triangle. $a_i = C(n-1, i), b_i = C(n, i)$, $c_i = C(n+1, i)$ \\ \\
3. Using the Skill $<$Combination$>$, $\sum_{i = 0}^{2006} \frac{b_i}{c_i} = \sum_{i = 0}^{2006} \frac{\binom{2006}{i}}{\binom{2007}{i}} = \sum_{i = 0}^{2006} \frac{2007 - i}{2007} = 2007 - \frac{2006 \cdot 2007}{2 \cdot 2007} = 2007-(2006*2007)/(2*2007) = 2007-1003 = 1004$ \\ \\
4. Using the Skill $<$Combination$>$, $\sum_{i = 0}^{2005} \frac{a_i}{b_i} = \sum_{i = 0}^{2005} \frac{\binom{2005}{i}}{\binom{2006}{i}} = \sum_{i = 0}^{2005} \frac{2006 - i}{2006} = 2006 - \frac{2005 \cdot 2006}{2 \cdot 2006} = 2006-(2005*2006)/(2*2006) = 2006-1002.5 = 1003.5$ \\ \\ 
5. Using the Skill $<$Sub$>$, $[\sum_{i = 0}^{2006} \frac{b_i}{c_i} - \sum_{i = 0}^{2005} \frac{a_i}{b_i}] = 1004-1003.5=0.5$ \\ \\
6. The answer is 0.5. 
\end{tcolorbox}
\noindent\begin{minipage}{\textwidth}   
\captionof{figure}{An exemplar of skill composition in Skills-in-Context prompting for MATH problems.} \label{Tab:compose_math_skill}
\end{minipage}

\newpage

\begin{tcolorbox}[title = {Skills for SACN}, colback = Apricot!25!white, colframe = BrickRed!75!black] 
Skill $<$command\_parse$>$: Parse a simple command into structured output. 
 \\
For example, ‘Walk’ is parsed to ‘WALK’,‘Look’ is parsed to ‘LOOK’,‘Run’ is parsed to ‘RUN’,‘Jump’ is parsed to ‘JUMP’’
‘turn left’ is parsed to ‘LTURN’’,‘turn right’ is parsed to ‘RTURN’’,
‘Turn opposite left’ is parsed to ‘LTURN LTURN’,
‘Turn opposite right’  is parsed to 'RTURN RTURN' \\ \\

Skill $<$decompose\_commands$>$: Find the max number in a list. \\
For example, ‘Look right’ can be decomposed to ‘look’ and ‘turn right’,‘Look twice’ can be decomposed to ‘look’ and ‘look’,‘Look right after look twice’ can be decompose to ‘look right’ and ‘look twice’,‘Jump opposite right thrice and walk’ can be decomposed to ‘jump opposite right thrice’, ‘walk’ \\ \\

\end{tcolorbox}
\noindent\begin{minipage}{\textwidth}
\captionof{figure}{The skills in Skills-in-Context prompt for the task of SACN.} \label{Tab:scan_skill}
\end{minipage}

\newpage

\begin{tcolorbox}[title = {An Example of Skill Composition for SCAN}, colback = Apricot!25!white, colframe = BrickRed!75!black] 
\textbf{Example}:  run and look twice. \\ \\
 \quad

\textbf{Answer}:

1. Using the Skill $<$decompose\_commands$>$, ‘run and look twice’ can be decomposed to ‘run’, and ‘look twice’. \\ \\
2.  For ‘run’, using the Skill $<$command\_parse$>$, ‘run’ is parsed to ‘RUN’
\\ \\
3. For ‘look twice’, using the Skill $<$decompose\_commands$>$, ‘look twice’ can be decomposed to ‘look’ and ‘look’: \\

 \qquad   i. For ‘look’, using the  Skill $<$command\_parse$>$, ‘look’ is parsed to ‘LOOK’

  \qquad  ii. For ‘look’, using the  Skill $<$command\_parse$>$, ‘look’ is parsed to ‘LOOK’
\\ \\
4. The output is ‘RUN’+’LOOK’+’LOOK’ = ‘RUN’ + ‘LOOK’*2

\end{tcolorbox}
\noindent\begin{minipage}{\textwidth}   
\captionof{figure}{An exemplar of skill composition in Skills-in-Context prompting for SCAN.} \label{Tab:compose_scan_skill}
\end{minipage}

\newpage

\begin{tcolorbox}[title = {Last Letter Concatenation}, colback = Apricot!25!white, colframe = BrickRed!75!black] 
\textbf{Question}: Take the last letters of the words in 'than, appropriate, sees, daisy, disturbed, espn, stable, double, luther, shut' and concatenate them.
\tcblower
\textbf{Model Output}:

\quad

1. Using the Skill $<$words\_to\_list$>$, put the asked words, 'than, appropriate, sees, daisy, disturbed, espn, stable, double, luther, shut', to a list.  D=['than', 'appropriate', 'sees', 'daisy', 'disturbed', 'espn', 'stable', 'double', 'luther', 'shut'] \\ \\
2. Get the last letter of each word in the list D=['than', 'appropriate', 'sees', 'daisy', 'disturbed', 'espn', 'stable', 'double', 'luther', 'shut'] to a new list R=[]: 

\quad

   \qquad i. Using the Skill $<$last\_letter$>$, the last letter of D[0]='than' is 'n'. R=[n] 
   
   \qquad ii. Using the Skill $<$last\_letter$>$, the last letter of D[1]='appropriate' is 'e'. R=[n, e] 
   \qquad iii. Using the Skill $<$last\_letter$>$, the last letter of D[2]='sees' is 's'. R=[n, e, s] 
   
   \qquad iv. Using the Skill $<$last\_letter$>$, the last letter of D[3]='daisy' is 'y'. R=[n, e, s, y] 
   
   \qquad v. Using the Skill $<$last\_letter$>$, the last letter of D[4]='disturbed' is 'd'. R=[n, e, s, y, d] 
   
   \qquad vi. Using the Skill $<$last\_letter$>$, the last letter of D[5]='espn' is 'n'. R=[n, e, s, y, d, n] 
   
   \qquad vii. Using the Skill $<$last\_letter$>$, the last letter of D[6]='stable' is 'e'. R=[n, e, s, y, d, n, e] 
   
   \qquad viii. Using the Skill $<$last\_letter$>$, the last letter of D[7]='double' is 'e'. R=[n, e, s, y, d, n, e, e] 
   
   \qquad ix. Using the Skill $<$last\_letter$>$, the last letter of D[8]='luther' is 'r'. R=[n, e, s, y, d, n, e, e, r] 
   
   \qquad x. Using the Skill $<$last\_letter$>$, the last letter of D[9]='shut' is 't'. R=[n, e, s, y, d, n, e, e, r, t] 

\quad

3. R=[n, e, s, y, d, n, e, e, r, t]. The answer is 'n e s y d n e e r t'. 
\end{tcolorbox}
\noindent\begin{minipage}{\textwidth}
\captionof{figure}{An example of the generated answer on last-letter-concatenation task using ChatGPT with our Skills-in-Context prompting.} \label{Tab:example_last_letter_skill}
\end{minipage}

\newpage

\begin{tcolorbox}[title = {Addition}, colback = Apricot!25!white, colframe = BrickRed!75!black] 
\textbf{Question}: Calculate 165789+876534
\tcblower
\textbf{Model Output}:

\quad

1. Using Skill $<$extract\_digits$>$ and Skill $<$list\_length$>$, extract the digits in 165789 to DM=[1,6,5,7,8,9].  len(DM)=6. Extract the digits in 876534 to DN=[8,7,6,5,3,4]. len(DN)=6. \\ \\
2. len(DM)=6=len(DN)=6. Align the two lists: DM=[1,6,5,7,8,9], DN=[8,7,6,5,3,4]. \\ \\
3. Using Skill $<$add\_two\_single\_digit\_number$>$, add every number in DM=[1,6,5,7,8,9] to the corresponding  number in DN =[8,7,6,5,3,4] to form a list, DS=[1+8,6+7,5+6,7+5,8+3,9+4]=[9,13,11,12,11,13]. \\ \\
4. Append a 0 to the beginning of DS: DS=[0,9,13,11,12,11,13]. Construct a list R=[0,0,0,0,0,0,0] with len(DS)=7 zeros. \\ \\
5. DS=[0,9,13,11,12,11,13] (DS[0]=0, DS[1]=9, DS[2]=13, DS[3]=11, DS[4]=12, DS[5]=11, DS[6]=13), starting from DS[6] to DS[1] (DS[6]=13, DS[5]=11, DS[4]=12, DS[3]=11, DS[2]=13, DS[1]=9): 

\quad

    \qquad i. DS[6]=13, R[6]=0, R[6]=R[6]+DS[6]=0+13=13. Based on Skill $<$compare\_10$>$, 13$>$10, R[6]=13$>$10, so R[5]=1 and R[6]=13-10=3 by Skill $<$sub\_10$>$. R=[R[0],R[1],R[2],R[3],R[4],R[5],R[6]]=[0,0,0,0,0,1,3]. 
    
    \qquad ii. DS[5]=11, R[5]=1, R[5]=R[5]+DS[5]=1+11=12. Based on Skill $<$compare\_10$>$, 12$>$10, R[5]=12$>$10, so R[4]=1 and R[5]=12-10=2 by Skill $<$sub\_10$>$. R=[R[0],R[1],R[2],R[3],R[4],R[5],R[6]]=[0,0,0,0,1,2,3].

    \qquad iii. DS[4]=12, R[4]=1, R[4]=R[4]+DS[4]=1+12=13. Based on Skill $<$compare\_10$>$, 13$>$10, R[4]=13$>$10, so R[3]=1 and R[4]=13-10=3 by Skill $<$sub\_10$>$. R=[R[0],R[1],R[2],R[3],R[4],R[5],R[6]]=[0,0,0,1,3,2,3]. 
    
    \qquad iv. DS[3]=11, R[3]=1, R[3]=R[3]+DS[3]=1+11=12. Based on Skill $<$compare\_10$>$, 12$>$10, R[3]=12$>$10, so R[2]=1 and R[3]=12-10=2 by Skill $<$sub\_10$>$. R=[R[0],R[1],R[2],R[3],R[4],R[5],R[6]]=[0,0,1,2,3,2,3]. 
    
    \qquad v. DS[2]=13, R[2]=1, R[2]=R[2]+DS[2]=1+13=14. Based on Skill $<$compare\_10$>$, 14$>$10, R[2]=14$>$10, so R[1]=1 and R[2]=14-10=4 by Skill $<$sub\_10$>$. R=[R[0],R[1],R[2],R[3],R[4],R[5],R[6]]=[0,1,4,2,3,2,3]. 
    
    \qquad vi. DS[1]=9, R[1]=1, R[1]=R[1]+DS[1]=1+9=10. Based on Skill $<$compare\_10$>$, 10=10, R[1]=10=10, so R[0]=1 and R[1]=10-10=0 by Skill $<$sub\_10$>$. R=[R[0],R[1],R[2],R[3],R[4],R[5],R[6]]=[1,0,4,2,3,2,3]. \\ 
6. R=[1,0,4,2,3,2,3]. The answer is 1042323. 
\end{tcolorbox}
\noindent\begin{minipage}{\textwidth}
\captionof{figure}{An example of the generated answer on the addition task using ChatGPT with Skills-in-Context prompting.} \label{Tab:example_simple_add_skill}
\end{minipage}

\newpage

\begin{tcolorbox}[title = {Multiplication}, colback = Apricot!25!white, colframe = BrickRed!75!black] 
\textbf{Question}: Calculate 1467*7651
\tcblower
\textbf{Model Output}:

\quad

1. Using Skill $<$extract\_digits$>$ and Skill $<$list\_length$>$, extract the digits in 1467 to DM=[1,4,6,7]. len(DM)=4. Extract the digits in 7651 to DN=[7,6,5,1]. len(DN)=4. \\  \\
2. Add 0,1,len(DM)-1=3 zeros to the end of every number in DM=[1,4,6,7] according to the position of the number in DM: DMO=[1*1000,4*100,6*10,7*1]=[1000,400,60,7]. \\ \\
3. Add 0,len(DN)-1=3 zeros to the end of every number in DN=[7,6,5,1] according to the position of the number in DN: DNO=[7*1000,6*100,5*10,1*1]=[7000,600,50,1]. \\ \\
4. Using Skill $<$mul\_two\_single\_digit\_number$>$, multiple every number in DMO=[1000,400,60,7] with every number in DNO=[7000,600,50,1] and get R=[1000*7000,1000*600,1000*50,1000*1, 400*7000,400*600,400*50,400*1,60*7000,60*600,60*50,60*1,7*7000,7*600,7*50,7*1]=  [7000000,600000,50000,1000,2800000,240000,20000,400,420000,36000,3000,60,\\49000,4200,350,7]. \\ \\
5. Using Skill $<$add\_multiple\_numbers$>$, add all the numbers in R=[7000000,600000,50000,1000, 2800000,240000,20000,400,420000,36000,3000,60,\\49000,4200,350,7], 7000000+600000+50000+1000+2800000+240000+20000+400+420000\\+36000+3000+60+49000+4200+350+7: 

\quad 

    \qquad i. Add the first two numbers: r1=7000000+600000=7600000. 
    
    \qquad ii. Add the third number 50000 to r1=7600000: r2=7600000+50000=7650000. 
    
    \qquad iii. Add the fourth number 1000 to r2=7650000: r3=7650000+1000=7651000. 
    
    \qquad iv. Add the fifth number 2800000 to r3=7651000: r4=7651000+2800000=10451000. 
    
    \qquad v. Add the sixth number 240000 to r4=10451000: r5=10451000+240000=10691000. 
    
    \qquad vi. Add the seventh number 20000 to r5=10691000: r6=10691000+20000=10711000. 
    
    \qquad vii. Add the eighth number 400 to r6=10711000: r7=10711000+400=10711400. 
    
    \qquad viii. Add the ninth number 420000 to r7=10711400: r8=10711400+420000=11131400. 
    
    \qquad ix. Add the tenth number 36000 to r8=11131400: r9=11131400+36000=11167400. 
    
    \qquad x. Add the eleventh number 3000 to r9=11167400: r10=11167400+3000=11170400. 
    
    \qquad xi. Add the twelfth number 60 to r10=11170400: r11=11170400+60=11170460. 
    
    \qquad xii. Add the thirteenth number 49000 to r11=11170460: r12=11170460+49000=11219460. 
    
    \qquad xiii. Add the fourteenth number 4200 to r12=11219460: r13=11219460+4200=11223660. 
    
    \qquad xiv. Add the fifteenth number 350 to r13=11223660: r14=11223660+350=11224010.
    
    \qquad xv. Add the sixteenth number 7 to r14=11224010: r15=11224010+7=11224017. \\ \\
6. So the answer is 11224017 
\end{tcolorbox}
\noindent\begin{minipage}{\textwidth}
\captionof{figure}{An example of the generated answer on the multiplication task using ChatGPT with Skills-in-Context prompting.} \label{Tab:example_simple_mul_skill}
\end{minipage}

\newpage

\begin{tcolorbox}[title = {CommaQA-E}, colback = Apricot!25!white, colframe = BrickRed!75!black] 
\textbf{Question}: 

\quad

movie: Skob ; directed by: Haldron. movie: Riften ; directed by: Conanopeia. movie: Tetroxidine ; directed by: Conanopeia. movie: Sheepcrest ; directed by: Flumph. movie: Sheepcrest ; award: Mariskenna. movie: Skob ; award: Jubeus. movie: Tetroxidine ; award: Handt. movie: Riften ; awarded: Handt. movie: Tetroxidine ; written by: Conanopeia. movie: Riften ; written by: Conanopeia. movie: Riften ; writer: Flumph. movie: Tetroxidine ; written by: Haldron. movie: Skob ; written by: Haldron. movie: Sheepcrest ; writer: Flumph. movie: Skob ; year: 1962. movie: Tetroxidine ; release year: 1997. movie: Riften ; release year: 1972. movie: Sheepcrest ; release year: 1997. Flumph was an actor in the movie Riften. Flumph acted in the movie Sheepcrest. Conanopeia was an actor in the movie Riften. Conanopeia acted in the movie Tetroxidine. Conanopeia acted in the movie Sheepcrest. Haldron acted in the movie Tetroxidine. Haldron acted in the movie Skob. Haldron was born in the year 1939. Flumph was born in the year 1938. Conanopeia was born in 1938. Haldron is from the country of Poquet. Conanopeia grew up in the nation of Poquet. Flumph grew up in the nation of Stridery. Diaqum was awarded to Flumph. Lidus was awarded to Conanopeia. Haldron won the Lidus award. Conanopeia was one of the producers of the movie Tetroxidine. Flumph was one of the producers of the movie Sheepcrest. Flumph produced the movie Skob with others. Haldron was one of the producers of the movie Sheepcrest. Haldron was one of the producers of the movie Riften. Haldron produced the movie Skob with others. Conanopeia was one of the producers of the movie Riften. \\ \\

Q: What movies have the people from Poquet written?

\tcblower
\textbf{Model Output}:

\quad

1. Using Skill $<$decompose\_qa$>$, decompose the question "What movies have the people from Poquet written?" into the following sub-questions: 

\quad

  \qquad  Q1: Who is from the country Poquet? 
  
   \qquad Q2: Which movies did [A1] write? 

\quad
   
2. Using Skill $<$answer\_simple\_question$>$, answer Q1: Who is from the country Poquet? 

Conanopeia grew up in the nation of Poquet. Haldron is from the country of Poquet. A1=["Conanopeia", "Haldron"] \\
3. A1=["Conanopeia", "Haldron"]. Using Skill $<$answer\_simple\_question$>$, answer Q2: Which movies did [A1] write? 

\quad

   \qquad  i. A1[0]="Conanopeia", answer Q2.1: Which movies did Conanopeia write? 

   \qquad \qquad  A2.1: movie: Tetroxidine ; written by: Conanopeia. movie: Riften ; written by: Conanopeia.   A2.1=["Tetroxidine", "Riften"] 
   
    \qquad ii. A1[1]="Haldron", answer Q2.2: Which movies did Haldron write? 
    
    \qquad \qquad  A2.2: movie: Skob ; written by: Haldron. A2.2=["Skob"] 
    
   \qquad  iii. So A2=A2.1+A2.2=["Tetroxidine", "Riften", "Skob"] \\ \\
4. The answer is ["Tetroxidine", "Riften", "Skob"] \\  
\end{tcolorbox}
\noindent\begin{minipage}{\textwidth}
\captionof{figure}{An example of the generated answer on the CommaQA-E task using ChatGPT with our Skills-in-Context prompting.} \label{Tab:example_compose_qa_skill}
\end{minipage}

\newpage

\begin{tcolorbox}[title = {DP}, colback = Apricot!25!white, colframe = BrickRed!75!black] 
\textbf{Question}:  Given a sequence of integers, find a subsequence with the highest sum, such that no two numbers in the subsequence are adjacent in the original sequence.  \\ 
A=[1, -5, 3, -4, -1, -1, 2, 2]. \\

\tcblower
\textbf{Model Output}:

\quad

1. A=[1, -5, 3, -4, -1, -1, 2, 2] (A[0]=1,A[1]=-5,A[2]=3,A[3]=-4,A[4]=-1,A[5]=-1,A[6]=2,A[7]=2). Using Skill $<$list\_length$>$, len(A)=8. Construct a list D=[0,0,0,0,0,0,0,0] with the len(A)=8 zeros. \\ \\
2. Update the last element in D. A[7]=2. Using Skill $<$find\_max$>$, max(2,0)=2. D[7]=2 and D=[0,0,0,0,0,0,0,2].  \\ \\
3. Update the second last element in D. A[7]=2, A[6]=2. Using Skill $<$find\_max$>$,  max(2,2,0)=2. D[6]=2 and D=[0,0,0,0,0,0,2,2]. \\ \\
4. Starting from the third last (len(A)-2=6) element in D, update each element in D based on A=[1, -5, 3, -4, -1, -1, 2, 2] (A[0]=1,A[1]=-5,A[2]=3,A[3]=-4, A[4]=-1,A[5]=-1,A[6]=2,A[7]=2): 

\quad

 \qquad i. D[7]=2, D[6]=2, A[5]=-1.  Using Skill $<$add$>$, A[5]+D[7]=-1+2=1. Using Skill $<$find\_max$>$, max(D[6],A[5]+D[7],0)=max(2,1,0)=2. D[5]=2 and  D=[0,0,0,0,0,2,2,2]. 
 
   \qquad  ii. D[6]=2, D[5]=2, A[4]=-1.  Using Skill $<$add$>$, A[4]+D[6]=-1+2=1.  Using Skill $<$find\_max$>$, max(D[5],A[4]+D[6],0)=max(2,1,0)=2. D[4]=2 and  D=[0,0,0,0,2,2,2,2]. 
   
   \qquad  iii. D[5]=2, D[4]=2, A[3]=-4.  Using Skill $<$add$>$, A[3]+D[5]=-4+2=-2. Using Skill $<$find\_max$>$, max(D[4],A[3]+D[5],0)=max(2,-2,0)=2. D[3]=2 and  D=[0,0,0,2,2,2,2,2]. 
   
   \qquad  iv. D[4]=2, D[3]=2, A[2]=3.  Using Skill $<$add$>$, A[2]+D[4]=3+2=5.  Using Skill $<$find\_max$>$, max(D[3],A[2]+D[4],0)=max(2,5,0)=5. D[2]=5 and  D=[0,0,5,2,2,2,2,2]. 
   
   \qquad  v. D[3]=2, D[2]=5, A[1]=-5.  Using Skill $<$add$>$, A[1]+D[3]=-5+2=-3.  Using Skill $<$find\_max$>$, max(D[2],A[1]+D[3],0)=max(5,-3,0)=5. D[1]=5 and  D=[0,5,5,2,2,2,2,2]. 
   
   \qquad  vi. D[2]=5, D[1]=5, A[0]=1.  Using Skill $<$add$>$, A[0]+D[2]=1+5=6.  Using Skill $<$find\_max$>$, max(D[1],A[0]+D[2],0)=max(5,6,0)=6. D[0]=6 and  D=[6,5,5,2,2,2,2,2]. \\ \\
5. D=[6,5,5,2,2,2,2,2]. The highest sum is D[0]=6. 
\end{tcolorbox}
\noindent\begin{minipage}{\textwidth}
\captionof{figure}{An example of the generated answer on the dynamic programming task using ChatGPT with our Skills-in-Context prompting.} \label{Tab:example_compose_dp_skill}
\end{minipage}

\newpage

\begin{tcolorbox}[title = {GSM8K}, colback = Apricot!25!white, colframe = BrickRed!75!black] 
\textbf{Question}:  The girls are trying to raise money for a carnival. Kim raises \$320 more than Alexandra, who raises \$430, and Maryam raises \$400 more than Sarah, who raises \$300. How much money, in dollars, did they all raise in total?

\tcblower
\textbf{Model Output}:

\quad

1. Alexandra raises \$430. \\ \\
2. Kim raises \$320 more than Alexandra. Using the Skill $<$add$>$, Kim raises \$430+\$320=\$750. \\ \\
3. Sarah raises \$300. \\ \\
4. Maryam raises \$400 more than Sarah. Using the Skill $<$add$>$, Maryam raises \$300+\$400=\$700. \\ \\
5. Using the Skill $<$add$>$, they all raise \$430+\$750+\$300+\$700=\$2180 in total. \\ \\
6. The answer is 2180.
\end{tcolorbox}
\noindent\begin{minipage}{\textwidth}
\captionof{figure}{An example of the generated answer on the GSM8K task using ChatGPT with Skills-in-Context prompting.} \label{Tab:example_compose_gsm8k_skill}
\end{minipage}

\newpage

\begin{tcolorbox}[title = {GSM8K}, colback = Apricot!25!white, colframe = BrickRed!75!black] 
\textbf{Question}: It's strawberry-picking time on Grandma Concetta's farm.  Tony can pick 6 quarts of strawberries per hour, while Bobby picks one less quart of strawberries per hour than Tony.  Kathy can pick twice as many strawberries per hour as Bobby, and Ricky picks two fewer quarts of strawberries per hour than does Kathy.  In total, how many quarts of strawberries can Tony, Bobby, Ricky, and Kathy pick per hour on  Grandma Concetta's farm? 
\tcblower
\textbf{Model Output}:

1. Tony can pick 6 quarts of strawberries per hour. \\ \\
2. Bobby picks one less quart of strawberries per hour than Tony. Using the Skill $<$sub$>$, Bobby picks 6-1=5 quarts of strawberries per hour. \\ \\
3. Kathy can pick twice as many strawberries per hour as Bobby. Using the Skill $<$mul$>$, Kathy picks 5*2=10 quarts of strawberries per hour. \\ \\
4. Ricky picks two fewer quarts of strawberries per hour than does Kathy.  Using the Skill $<$sub$>$, Ricky picks 10-2=8 quarts of strawberries per hour. \\ \\
5. In total, Tony, Bobby, Ricky, and Kathy can pick 6+5+10+8 quarts of strawberries per hour. Using the Skill \mybox[GreenYellow]{$<$add\_multiple\_numbers$>$}: 

\quad

   \qquad i. Add the first two numbers using Skill $<$add$>$: r1=6+5=11. 
   
     \qquad ii. Add the third number 10 to r1=11 using Skill $<$add$>$: r2=11+10=21. 
     
    \qquad  iii. Add the fourth number 8 to r2=21 using Skill $<$add$>$: r3=21+8=29. 

\quad
    
6. So the answer is 29. 
\end{tcolorbox}
\noindent\begin{minipage}{\textwidth}
\captionof{figure}{An example of the generated answer on the GSM8K task with our Skills-in-Context prompting, where $<$add\_multiple\_numbers$>$ is included as a basic skill in the SKiC prompting context (see Table \ref{Tab:gsm8k_skill}) but is not demonstrated in any given exemplar to show how to use it. LLMs automatically figure out how to use such skills in an innovative composition to solve an unseen complex problem.} \label{Tab:example_compose_gsm8k_skill_1}
\end{minipage}

\newpage

\begin{tcolorbox}[title = {GSM8K}, colback = Apricot!25!white, colframe = BrickRed!75!black] 
\textbf{Question}: A merchant wants to make a choice of purchase between 2 purchase plans: jewelry worth \$5,000 or electronic gadgets worth \$8,000. His financial advisor speculates that the jewelry market will go up 2.5\% while the electronic gadgets market will rise 1.2\% within the same month. If the merchant is looking to maximize profit at the end of this month by making a choice, how much profit would this be?
\tcblower
\textbf{Model Output}:

\quad

1. If the merchant buys jewelry worth \$5,000 and the jewelry market goes up 2.5\%, using the Skill $<$mul$>$, the value of the jewelry will increase by \$5,000*2.5\%= \$125.  Using Skill $<$add$>$, the value of the jewelry will be \$5,000+\$125=\$5125. \\ \\
2. If the merchant buys electronic gadgets worth \$8,000 and the electronic gadgets market goes up 1.2\%, using the Skill $<$mul$>$, the value of the electronic gadgets will increase by \$8,000*1.2\%= \$96.  Using Skill $<$add$>$, the value of the electronic gadgets will be \$8,000+\$96=\$8096. \\ \\
3. The merchant wants to maximize profit. Using the Skill  $<$sub$>$,  the profit from buying jewelry will be \$5125-\$5000=\$125.  The profit from buying electronic gadgets will be \$8096-\$8000=\$96. \\ \\
4. Using the Skill \mybox[GreenYellow]{$<$compare$>$}, \$125$>$\$96, so the merchant should buy jewelry  to maximize profit. The profit will be \$125. \\ \\
5. The answer is 125.

\end{tcolorbox}
\noindent\begin{minipage}{\textwidth}
\captionof{figure}{An example of the generated answer on the GSM8K task with our Skills-in-Context prompting, where the skill $<$compare$>$ are neither included in the SKiC prompting context (see Table \ref{Tab:gsm8k_skill}) nor used in any given exemplars. LLMs utilize the skills pre-existing in their pre-trained knowledge to solve the problem.} \label{Tab:example_compose_gsm8k_skill_2}
\end{minipage}

\newpage

\begin{tcolorbox}[title = {GSM8K}, colback = Apricot!25!white, colframe = BrickRed!75!black] 
\textbf{Question}: Mr. Jackson’s fourth-grade class has 27 students. He wants to give each student 2 glue sticks. The glue sticks  come in packs of 8. How many packs will Mr. Jackson need to buy so every student can have 2 glue sticks, assuming he can only buy whole packs and he expects to have some extra glue sticks left over?
\tcblower
\textbf{Model Output}:

\quad

1. Mr. Jackson’s fourth-grade class has 27 students and he wants to give each student 2 glue sticks. Using the Skill $<$mul$>$, Mr. Jackson needs 27*2=54 glue sticks. \\ \\
2. The glue sticks come in packs of 8. Using the Skill $<$div$>$, Mr. Jackson needs 54/8=6.75  packs of glue sticks. \\ \\
3. Mr. Jackson can only buy whole packs. Using the Skill \mybox[GreenYellow]{$<$round$>$},  Mr. Jackson needs to buy 7 packs of glue sticks.  \\ \\
4. The answer is 7.
\end{tcolorbox}
\noindent\begin{minipage}{\textwidth}
\captionof{figure}{An example of the generated answer on the GSM8K task with our Skills-in-Context prompting, where the skill $<$round$>$ are neither included in the original SKiC prompting context (see Table \ref{Tab:gsm8k_skill}) nor used in any given exemplars. LLMs utilize the skills pre-existing in their pre-trained knowledge to solve the problem.} \label{Tab:example_compose_gsm8k_skill_3}
\end{minipage}

\newpage

\begin{tcolorbox}[title = {MATH}, colback = Apricot!25!white, colframe = BrickRed!75!black] 
\textbf{Question}: What is the smallest possible average of four distinct positive even integers?
\tcblower
\textbf{Model Output}:

\quad

1. The smallest possible average of four distinct positive even integers is achieved when the four integers  are the smallest possible. \\ \\
2. The smallest four distinct positive even integers are 2, 4, 6, and 8. \\ \\
3. Using the Skill \mybox[GreenYellow]{$<$Average$>$}, the average of these four numbers is (2+4+6+8)/4=20/4=5. \\ \\
4. The answer is 5. 
\end{tcolorbox}
\noindent\begin{minipage}{\textwidth}
\captionof{figure}{An example of the generated answer on the MATH task with our Skills-in-Context prompting, where the skill $<$Average$>$  are neither included in the original SKiC prompting context (see Table \ref{Tab:math_skill}) nor used in any given exemplars. LLMs(GPT4) utilize the skills pre-existing in their pre-trained knowledge to solve the problem.} \label{Tab:example_compose_math_skill_1}
\end{minipage}

\newpage

\begin{tcolorbox}[title = {MATH}, colback = Apricot!25!white, colframe = BrickRed!75!black] 
\textbf{Question}: How many numbers are in the list $2008, 2003, 1998, \ldots, 8, 3?$
\tcblower
\textbf{Model Output}:

\quad

1. The list starts at 2008 and ends at 3, and the difference between consecutive terms is 5. \\ \\
2. Using the Skill \mybox[GreenYellow]{$<$Arithmetic Sequence$>$}, the n-th term of an arithmetic sequence can be found using the formula $a_n = a_1 + (n - 1)d$, where $a_n$ is the nth term, $a_1$ is the first term, $n$ is the term number, and $d$ is the common difference. \\ \\
3. Substitute $a_n$ with 3, $a_1$ with 2008, and d with $-5$ into the formula, $3 = 2008 + (n - 1)(-5)$. \\ \\
4. Using the Skill $<$Solve Equation$>$, move $2008$ to the other side, $(n - 1)(-5) = 3 - 2008$. Using the Skill $<$Sub$>$, $(n - 1)(-5) = -2005$. \\ \\
5. Using the Skill $<$Solve Equation$>$, move $*(-5)$ to the other side, $n - 1 = -2005 / -5$.  Using the Skill $<$Div$>$, $n - 1 = 401$. \\ \\
6. Using the Skill $<$Solve Equation$>$, move $-1$ to the other side, $n = 401 + 1$.  Using the Skill $<$Add$>$, $n = 402$. \\ \\
7. The answer is 402. \\
\end{tcolorbox}
\noindent\begin{minipage}{\textwidth}
\captionof{figure}{An example of the generated answer on the MATH task with our Skills-in-Context prompting, where the skill $<$Arithmetic Sequence$>$ are neither included in the original SKiC prompting context (see Table \ref{Tab:math_skill}) nor used in any given exemplars. LLMs(GPT4) utilize the skills pre-existing in their pre-trained knowledge to solve the problem.} \label{Tab:example_compose_math_skill_2}
\end{minipage}

\newpage

\begin{tcolorbox}[title = {MATH}, colback = Apricot!25!white, colframe = BrickRed!75!black] 
\textbf{Question}: Suppose $P$ is the point $(5,3)$ and $Q$ is the point $(-3,6)$. What is the midpoint of $\overline{PQ}$?
\tcblower
\textbf{Model Output}:

\quad

1. Using the Skill \mybox[GreenYellow]{$<$Midpoint Formula$>$}, the midpoint of two points $(x_1, y_1)$ and $(x_2, y_2)$ is given  by $\left(\frac{x_1+x_2}{2}, \frac{y_1+y_2}{2}\right)$. \\ \\
2. For points $P(5,3)$ and $Q(-3,6)$, the midpoint is $\left(\frac{5+(-3)}{2}, \frac{3+6}{2}\right)$. \\ \\
3. Using the Skill $<$Add$>$ and $<$Div$>$, the midpoint is $\left(\frac{2}{2}, \frac{9}{2}\right) = (1, 4.5)$. \\ \\
4. The answer is $(1, 4.5)$. \\
\end{tcolorbox}
\noindent\begin{minipage}{\textwidth}
\captionof{figure}{An example of the generated answer on the MATH task with our Skills-in-Context prompting, where the skill $<$Midpoint Formula$>$  are neither included in the original SKiC prompting context (see Table \ref{Tab:math_skill}) nor used in any given exemplars. LLMs(GPT4) utilize the skills pre-existing in their pre-trained knowledge to solve the problem.} \label{Tab:example_compose_math_skill_3}
\end{minipage}

\newpage

\begin{tcolorbox}[title = {MATH}, colback = Apricot!25!white, colframe = BrickRed!75!black] 
\textbf{Question}: If $\mathbf{a}$ and $\mathbf{b}$ are vectors such that $\|\mathbf{a}\| = 2,$ $\|\mathbf{b}\| = 7,$ and $\mathbf{a} \times \mathbf{b} = \begin{pmatrix} 3 \\ 2 \\ 6 \end{pmatrix}$, then find the smallest possible angle between $\mathbf{a}$ and $\mathbf{b}.$ 
\tcblower
\textbf{Model Output}:

\quad

1. Using the Skill \mybox[GreenYellow]{$<$Cross Product$>$}, the magnitude of the cross product of two vectors is equal to the  product of the magnitudes of the vectors and the sine of the angle between them. So, we have  $\|\mathbf{a} \times \mathbf{b}\| = \|\mathbf{a}\|\|\mathbf{b}\|\sin \theta$, where $\theta$ is the angle between $\mathbf{a}$ and $\mathbf{b}$. \\ \\
2. Using the Skill \mybox[GreenYellow]{$<$Vector Magnitude$>$}, the magnitude of the cross product $\|\mathbf{a} \times \mathbf{b}\|$ is $\sqrt{3^2 + 2^2 + 6^2} = \sqrt{49} = 7$. \\ \\
3. Substituting the given values, we get $7 = 2*7*\sin \theta$. Using the Skill $<$Div$>$ and $<$Mul$>$, we get $\sin \theta = 7/(2*7) = 1/2$. \\ \\
4. Using the Skill \mybox[GreenYellow]{$<$Inverse Trigonometric Functions$>$}, the angle whose sine is 1/2 is 30 degrees or 150 degrees. But we want the smallest possible angle, so $\theta = 30$ degrees. \\ \\
5. The answer is 30. 
\end{tcolorbox}
\noindent\begin{minipage}{\textwidth}
\captionof{figure}{An example of the generated answer on the MATH task with our Skills-in-Context prompting, where the skill  $<$Cross Product$>$, $<$Vector Magnitude$>$,$<$Inverse Trigonometric Functions$>$  are neither included in the original SKiC prompting context (see Table \ref{Tab:math_skill}) nor used in any given exemplars. LLMs(GPT4) utilize the skills pre-existing in their pre-trained knowledge to solve the problem.} \label{Tab:example_compose_math_skill_4}
\end{minipage}

\end{strip}
\newpage

\end{document}